\documentclass[10pt,twocolumn,letterpaper]{article}

 \usepackage[pagenumbers]{cvpr} % To force page numbers, e.g. for an arXiv version

\usepackage[dvipsnames]{xcolor}

\definecolor{cvprblue}{rgb}{0.21,0.49,0.74}
\usepackage[pagebackref,breaklinks,colorlinks,citecolor=cvprblue]{hyperref}
\usepackage[dvipsnames]{xcolor}
\usepackage{tikz}
\usetikzlibrary{positioning, shapes.geometric, fit, calc, backgrounds, external}
\pgfdeclarelayer{background}
\pgfdeclarelayer{foreground}
\pgfsetlayers{background,main,foreground}
\usepackage{tikzpagenodes} % Add this to your preamble
\usepackage{adjustbox}
\usepackage{float}
\usepackage[ruled,vlined, linesnumbered]{algorithm2e}
\usepackage{multirow}
\usepackage{booktabs}
\usepackage{pgfplots}
\usepackage{comment}
\usepackage{graphicx}
\usepackage{algorithmic}
\pgfplotsset{compat=1.18}

\usetikzlibrary{backgrounds}

\usepackage{pifont}% http://ctan.org/pkg/pifont
\newcommand{\cmark}{\ding{51}}%
\newcommand{\xmark}{\ding{55}}%

\def\paperID{17693} % *** Enter the Paper ID here
\def\confName{CVPR}
\def\confYear{2024}

\title{Decompose-and-Compose: A Compositional Approach to  Mitigating Spurious Correlation}

\author{\textbf{Fahimeh Hosseini Noohdani}\\
{\tt\small {\tt\small fhosseini@ce.sharif.edu}}
\and
\textbf{Parsa Hosseini}$^*$\\
{\tt\small parsa.hosseini@sharif.edu}
\and
\textbf{Aryan Yazdan Parast}$^*$\\
{\tt\small arian.yazdanparast@sharif.edu}
\and
\textbf{Hamidreza Yaghoubi Araghi}\\
{\tt\small hamidreza.yaghoubiaraghi@sharif.edu}
\and
\textbf{Mahdieh Soleymani Baghshah}\\
{\tt\small soleymani@sharif.edu}\\
\and
Sharif University of Technology\\
Tehran, Iran
}

\begin{document}
\maketitle
\def\thefootnote{*}\footnotetext{Equal contribution.}
\begin{abstract}
While standard Empirical Risk Minimization (ERM) training is proven effective for image classification on in-distribution data, it fails to perform well on out-of-distribution samples. 
One of the main sources of distribution shift for image classification is the compositional nature of images.
Specifically, in addition to the main object or component(s) determining the label, some other image components usually exist, which may 
lead to the shift of input distribution between train and test environments. More importantly, these components may have spurious correlations with the label.
To address this issue, we propose Decompose-and-Compose (DaC), which improves robustness to correlation shift by a compositional approach based on combining elements of images.  
Based on our observations, models trained with 
ERM usually highly attend to either the causal components or the components having a high spurious correlation with the label (especially in datapoints on which models have a high confidence). In fact, according to the amount of spurious correlation and the easiness of classification based on the causal or non-causal components, the model usually attends to one of these more (on samples with high confidence). Following this, we first try to identify the causal components of images using class activation maps of models trained with ERM. 
Afterwards, we intervene on images by combining them and retraining the model on the augmented data, including the counterfactual ones.
 This work proposes a group-balancing method by intervening on images without requiring group labels or information regarding the spurious features during training. The method has an overall better worst group accuracy compared to previous methods with the same amount of supervision on the group labels in correlation shift. Our code is available at \url{https://github.com/fhn98/DaC}.
\end{abstract}    
\section{Introduction}
\label{sec:intro}

\begin{figure*}[ht!]
\centering
  \noindent\begin{adjustbox}{max width=\textwidth}
\begin{tikzpicture}

\begin{axis}[
            at={(0,0)},
            ybar,
            bar width=0.4cm,
            width=8cm,
            height=3.5cm,
            legend style={at={(1,1.1)}, legend columns=-1,  font=\footnotesize,
            draw=none},
            y label style={at={(0.08,0.35)}, anchor=south west},
            symbolic x coords={0-0.25, 0.25-0.5, 0.5-0.75, 0.75-1},
            xtick=data,
            xlabel={Quantiles of Loss},
            x label style={at={(3.2cm,-0.15)}, anchor=south, font=\footnotesize},
            ymin=0,
            ymax=1,
            nodes near coords = {},
            tick label style={font=\fontsize{8}{10}\selectfont},
            axis x line*=bottom,
            axis y line*=left,
            major tick length=2pt,
            title={\footnotesize(a)},
        ]
        
        % Avg. C Score
        \addplot[fill=purple] coordinates {(0-0.25, 0.13) (0.25-0.5, 0.19) (0.5-0.75, 0.29) (0.75-1, 0.43)};
         \addlegendentry{$C$ Score}
        
         % Avg. S Score
         \addplot[fill=pink] coordinates {(0-0.25, 0.87) (0.25-0.5, 0.81) (0.5-0.75, 0.71) (0.75-1, 0.57)};
         \addlegendentry{$S$ Score}

        \end{axis}

    % Draw axis line
    \newcommand{\yshift}{3.1};\newcommand{\scale}{0.8};\newcommand{\xshift}{-0.8};
    
    \draw[-] (0+\xshift,0+\yshift) -- (10*\scale+\xshift,0+\yshift) node[right]{} ;
    
    % Draw tick marks and labels
    \foreach \x in {0,0.5,1}
        \draw (\x*\scale*10+\xshift,0.1+\yshift) -- (\x*\scale*10+\xshift,-0.1+\yshift) node[below] {\footnotesize \x};

    \node[] (q0d) at (0.75*\scale+\xshift,1.2+\yshift) {\includegraphics[width=1.2cm]{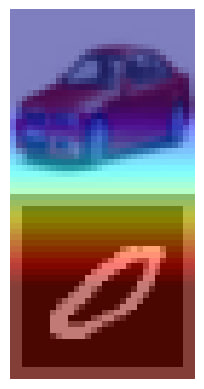}};

    \node[above=0.01cm of q0d, text width=1.5cm, align=center] (q0d_label) {\tiny 
    \vspace{-6pt}
    $C=0.12$ 
    \vspace{-6pt}
    $S=0.88$};

    \node[] (q1d) at (2.25*\scale+\xshift,1.2+\yshift) {\includegraphics[width=1.2cm]{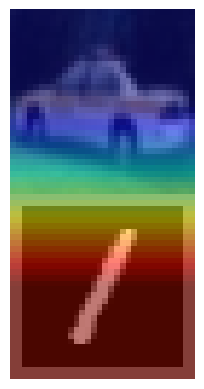}};

    \node[above=0.01cm of q1d, text width=1.5cm, align=center] (q1d_label) {\tiny 
    \vspace{-6pt}
    $C=0.12$ 
    \vspace{-6pt}
    $S=0.88$};

    \node (q2d) at (7.75*\scale+\xshift,1.2+\yshift) {\includegraphics[width=1.2cm]{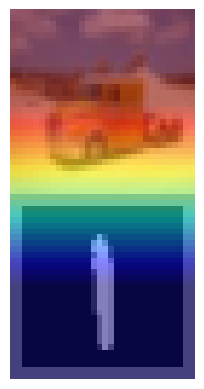}};

    \node[above=0.01cm of q2d, text width=1.5cm, align=center] (q2d_label) {\tiny
    \vspace{-6pt}
    $C=0.88$
    \vspace{-6pt}
    $S=0.12$};

    \node (q3d) at (9.25*\scale+\xshift,1.2+\yshift) {\includegraphics[width=1.2cm]{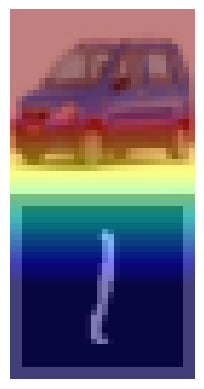}};

    \node[above=0.01cm of q3d, text width=1.5cm, align=center] (q3d_label) {\tiny
    \vspace{-6pt}
    $C=0.88$
    \vspace{-6pt}
    $S=0.12$};

    \node[align=center] at (3.2cm,-0.9) (b_label) {\footnotesize(b)};

    \newcommand{\xxshift}{9};

    \begin{axis}[
            at={(9cm,0)},
            ybar,
            bar width=0.4cm,
            width=8cm,
            height=3.5cm,
            legend style={at={(1,1.1)}, legend columns=-1,  font=\footnotesize,
            draw=none},
            y label style={at={(0.08,0.35)}, anchor=south west},
            symbolic x coords={0-0.25, 0.25-0.5, 0.5-0.75, 0.75-1},
            xtick=data,
            xlabel={Quantiles of Loss},
            x label style={at={(3.2cm,-0.15)}, anchor=south, font=\footnotesize},
            ymin=0,
            ymax=1,
            nodes near coords = {},
            tick label style={font=\fontsize{8}{10}\selectfont},
            axis x line*=bottom,
            axis y line*=left,
            major tick length=2pt,
            title={\footnotesize(c)},
        ]
        
        % Avg. C Score
        \addplot[fill=purple] coordinates {(0-0.25, 0.70) (0.25-0.5, 0.67) (0.5-0.75, 0.64) (0.75-1, 0.60)};
         \addlegendentry{$C$ Score}
        
         % Avg. S Score
         \addplot[fill=pink] coordinates {(0-0.25, 0.28) (0.25-0.5, 0.27) (0.5-0.75, 0.28) (0.75-1, 0.29)};
         \addlegendentry{$S$ Score}

        \end{axis}
        
    % Draw axis line
    \newcommand{\yshifta}{\yshift};\newcommand{\scalea}{0.8};\newcommand{\xshifta}{-0.8+\xxshift};
    
    \draw[-] (0+\xshifta,0+\yshifta) -- (10*\scalea+\xshifta,0+\yshifta) node[right]{} ;
    
    % Draw tick marks and labels
    \foreach \x in {0,0.5,1}
        \draw (\x*\scalea*10+\xshifta,0.1+\yshifta) -- (\x*\scalea*10+\xshifta,-0.1+\yshifta) node[below] {\footnotesize \x};

    \node[] (wb0) at (1*\scalea+\xshifta,0.9+\yshifta) {\includegraphics[width=1.5cm]{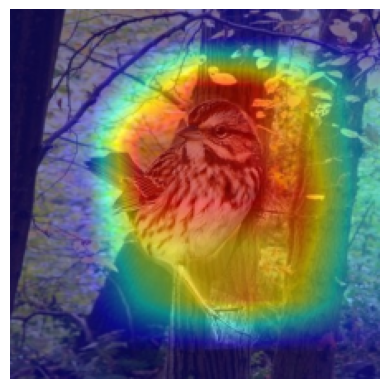}};

    \node[above=0.01cm of wb0, text width=1.5cm, align=center] (wb0_label) {\tiny 
    \vspace{-6pt}
    $C=0.79$ 
    \vspace{-6pt}
    $S=0.30$};

    \node[] (wb1) at (3*\scalea+\xshifta,0.9+\yshifta) {\includegraphics[width=1.5cm]{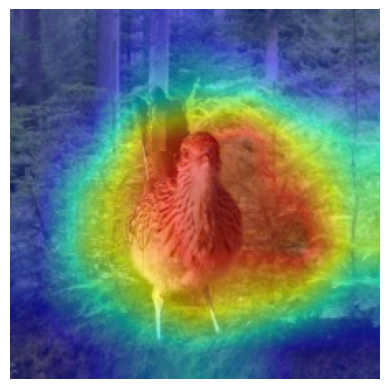}};

    \node[above=0.01cm of wb1, text width=1.5cm, align=center] (wb1_label) {\tiny 
    \vspace{-6pt}
    $C=0.79$ 
    \vspace{-6pt}
    $S=0.32$};

    \node (wb2) at (7*\scalea+\xshifta,0.9+\yshifta) {\includegraphics[width=1.5cm]{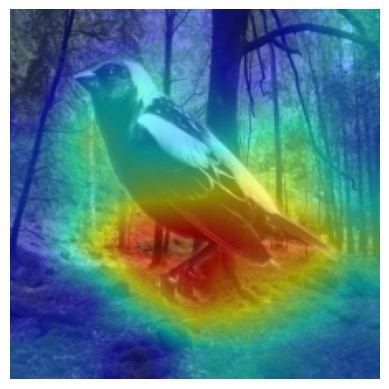}};

    \node[above=0.01cm of wb2, text width=1.5cm, align=center] (wb2_label) {\tiny
    \vspace{-6pt}
    $C=0.55$
    \vspace{-6pt}
    $S=0.32$};

    \node (wb3) at (9*\scalea+\xshifta,0.9+\yshifta) {\includegraphics[width=1.5cm]{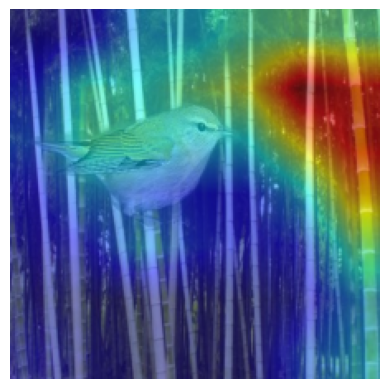}};

    \node[above=0.01cm of wb3, text width=1.5cm, align=center] (wb3_label) {\tiny
    \vspace{-6pt}
    $C=0.32$
    \vspace{-6pt}
    $S=0.28$};

    \node[align=center] at (12.2cm,-0.9) (b_label) {\footnotesize(d)};
\end{tikzpicture}
\end{adjustbox}
    \caption{Behaviour of a model trained with standard ERM in different datasets. Based on the easiness of inferring the label from the causal or non-causal parts across the whole dataset, the model attends more to one of them, this behaviour is more evident in samples on which the model has a low loss. 
    (a), (b) Average xGradCAM score of Cifar10 (causal) and MNIST (non-causal) pixels in four loss quantiles of the Dominoes training set. The model generally attends more to the \textit{non-causal} parts, and as the loss decreases, the \textit{non-causal} attention increases.
    (c), (d) Average xGradCAM score of foreground (causal) and background (non-causal) pixels in four loss quantiles of the Waterbirds training set. The model generally attends to the \textit{causal} parts, and as the loss decreases, the \textit{causal} attention increases.}

\label{fig:gradscore}
\end{figure*}
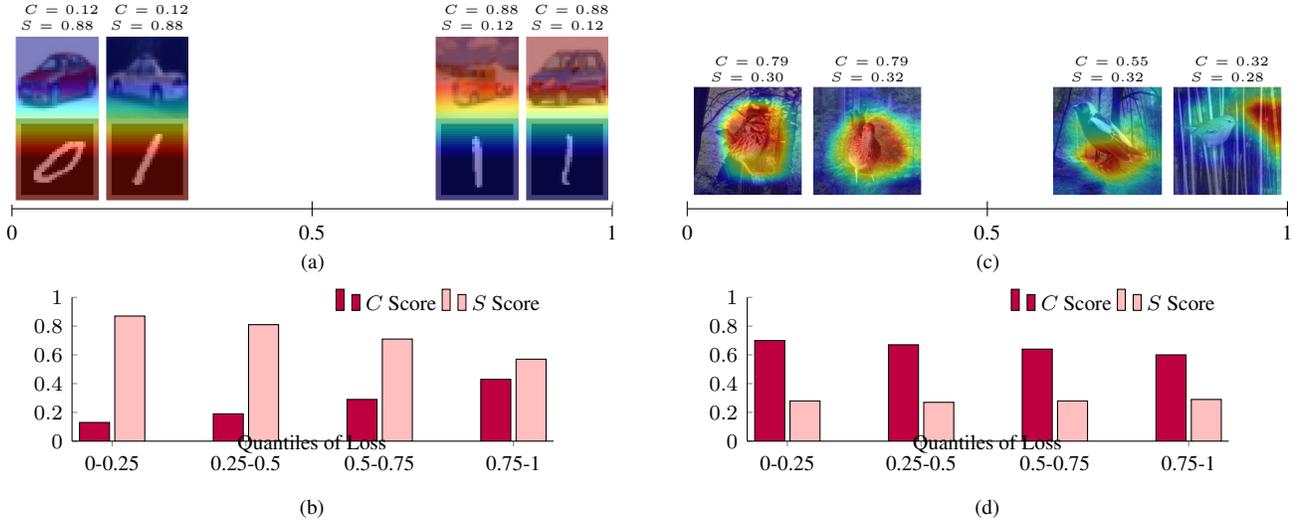

While deep neural networks are capable of superhuman performance, they still fail when faced with out-of-distribution (OOD) data~\cite{terra, group, irm}. Studies have shown that these models tend to make their predictions according to simple features that have a high correlation with the label, although these correlations are unstable across data distributions~\cite{bias, group, irm}. Relying on these spurious correlations instead of the stable ones causes the model to overfit on the training data and fail on OOD samples, for which those previous correlations do not hold. 

To tackle this challenge, methods based on invariant learning focus on learning representations that can be used to make invariant predictions across different environments, to make the trained model more robust to distribution shifts~\cite{irm, ReX, Fishr, PGI}.

Another line of work approaches this problem
by balancing minority/majority groups of data~\cite{LastLayer, SUBG, balancesagawa} to remove spurious correlation. Among these works, \cite{LastLayer} proposes DFR, which retrains the last layer of a model previously trained by ERM, with group-balanced data to make it robust to spurious correlation. 
Nonetheless, DFR requires group labels. On the other hand, methods like \cite{jtt, AFR, LFF} try to learn a robust model by upweighting samples that are less likely to contain spurious attributes, without access to group labels during training. Relying on the assumption that datapoints on which the model has a high loss are most probably from minority groups, most of these methods aim to place more emphasis on these samples. However, in these samples, the obscure core object may be the source of high loss (i.e., the target object itself cannot be easily classified), and overrating these samples may have some side effects.
To be more precise, we need to discover and analyze parts of images to make a more accurate decision.

Due to the compositional nature of images, the problem of correlation shift can be viewed through the lens of compositionality, as models fall into the trap of spurious correlation because they make their predictions based on non-causal components of images. 
This fine-grained perspective could lead to a more precise approach compared to methods that consider images as a whole.
 While the viewpoint of compositionality is essential to OOD generalization, especially when facing correlation shift, only a limited number of works have explored this problem from this perspective.
 As a recent work, \cite{maskfinetune} combines parts of different images and uses them for model distillation on the representation level. Nonetheless, they cannot label the combined images, as they could not determine whether the parts taken from images for combining are causal or non-causal parts. Additionally, they do not offer any evaluations on correlation shift benchmarks. Masktune \cite{Masktune} takes a step further and hypothesizes that in datasets exhibiting spurious correlation, the parts of an image with high attribution scores according to a model trained with ERM, are non-causal and misleading, and based on this assumption, masks these parts for finetuning the model. 

Inspired by this compositional viewpoint, we propose Decompose-and-Compose (DaC), a method for balancing groups by intervening on non-causal components of images and creating new ones. 
The same idea of intervening on images or using synthetic data as a means of group-balancing has been previously studied in a few works~\cite{ffr, discoverandcure}. However, unlike our method, which does not require any external aid during training, both these studies  
have a knowledge of the possible spurious attributes, and based on this knowledge, they create concept banks~\cite{discoverandcure} or intervene in images using generative models~\cite{ffr}.

\begin{figure*}[ht!]\label{scm}
\begin{subfigure}{.5\textwidth}
  \centering
  \includegraphics[width=.8\linewidth]{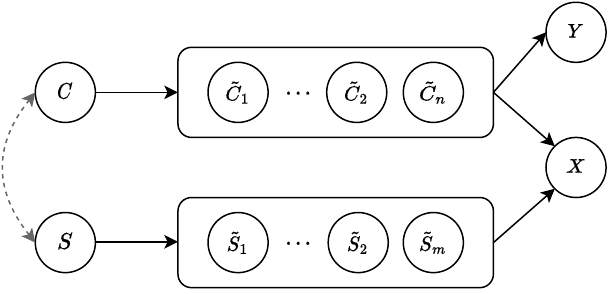}
  \caption{}
  \label{fig:scm_data}
\end{subfigure}%
\begin{subfigure}{.5\textwidth}
  \centering
  \includegraphics[width=.83\linewidth]{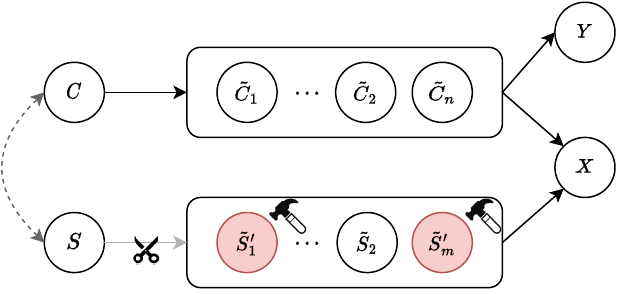}
  \caption{}
  \label{fig:scm_intervene}
\end{subfigure}
\caption{(a) Image as a composition of causal and non-causal components. (b) The edge between $S$ and $\tilde{S}$ can be removed by intervention on components in $\tilde{S}$. This removes the spurious correlation between $\Tilde{S}$ and $Y$.}
\end{figure*}

In this paper, we first analyze the behaviour of a model trained with ERM and utilize its attribution map on images to decompose them into causal and non-causal parts; then, based on the performance of the model trained with our method, on the validation set, we identify the causal parts.

More precisely, as opposed to MaskTune~\cite{Masktune}, which assumes that for a given model trained with standard ERM, the regions with high attribution scores are spurious ones, we show that a model trained with ERM usually focuses on either causal or non-causal parts of images, based on the easiness of predicting the label from them.
Gaining this knowledge about the causal parts enables us to intervene on datapoints from the majority groups to create new ones from under-represented groups as a means of group balancing.

The contributions of this work are as follows:
\begin{itemize}
\item We provide an analysis of the behaviour of models trained with standard ERM, especially on low-loss data.
\item Based on this analysis, we introduce a method for identifying the causal parts of an image.
\item We propose a method for combining images, to create new datapoints representing the minority groups, as a means for group balancing. 
\item Our proposed method performs better than previous baselines on well-known benchmarks in the literature. 
\end{itemize}

\section{Preliminaries}
\label{sec:preliminaries}
Spurious correlations that are perceivable by humans can be categorized into two groups: 1) Spurious parts: spurious correlations between other objects of the image and the label (e.g., spurious correlation of the background and the label in the Waterbirds~\cite{group} dataset) and 2) Spurious attributes: spurious correlations between some non-causal attributes of the object of interest and the label (e.g., spurious correlation between the colour of the digits and their label in the CMNIST~\cite{irm} dataset). In this section, we propose a method that provides robustness to the first type of spurious correlation, which is prevalent in most benchmarks. The proposed method mitigates correlation shifts by discovering non-causal parts of images and intervening on these parts.

\subsection{Problem Definition}
\label{sec:problemdefinition}
Consider a dataset $D_{tr} = \{(x^{(i)}, y^{(i)})\}_{i=1}^N$ for a classification problem. Each $x\in\mathcal{X}$ has \textit{core} features $c$, which are the cause of the label, and their correlation with the label is persistent across different environments. It has also \textit{spurious} features $s$. Based on combinations of core and spurious features, $(c,s)$, training samples can be partitioned into several groups. We consider the case when there is an imbalance between the size of groups in the training set. In this case, the group containing the majority of samples in a class is called the \textit{majority} group of that class, and the others are called the \textit{minority} groups. This imbalance induces a \textit{spurious correlation} between the spurious features and the label, i.e. the value of the spurious features and the label corresponding to a majority group are frequently seen together in the dataset. The proportion of samples in groups could be different in the test set, causing a \textit{correlation shift} between the training and the test sets. For instance, in the Waterbirds dataset~\cite{group},
in which the task is to determine whether each image shows a waterbird or a landbird,
the core and spurious features are the foreground and the background respectively. Waterbirds consists of four groups: waterbirds on water background, landbirds on land background, waterbirds on land background and landbirds on water background, with the first two being the majority groups.

Our objective is to train a classifier, denoted as $f$, that performs well across both the training and test distributions. This entails ensuring that $f$ exhibits strong performance not only on the majority but also on the minority groups.

\subsection{A Causality Viewpoint to Spurious Correlation}
\label{sec:causal}
To study the problem of spurious correlation from the perspective of causality, we model the data-generating process as the \underline{S}tructural \underline{C}ausal \underline{M}odel (SCM)~\cite{pearl} shown in \cref{fig:scm_data}. In this SCM, $C$ and $S$ indicate unobservable causal and non-causal variables, from which the observable causal and non-causal components $\tilde{C}$ and $\tilde{S}$ for an image are obtained. The final image $X$ is the output of $\psi(\tilde{C},\tilde{S})$, where $\psi(.,.)$ is a combining function. The label of the image is caused by $\tilde{C}$.

In the case of spurious correlation, a hidden confounder $E$, mostly referred to as the \textit{environment}~\cite{irm} or \textit{group}~\cite{group} variable in the literature, would be present in the SCM such that $S\leftarrow E\rightarrow C$. This creates the path $\tilde{S}\leftarrow S \leftarrow E \rightarrow C \rightarrow \tilde{C}\rightarrow {Y}$, which introduces a spurious correlation between $\tilde{S}$ and $Y$. $E$ is mainly sample selection bias. 

Whereas most previous studies on mitigating spurious correlation did not approach this problem from a causality perspective, methods based on group balancing, such as ~\cite{LastLayer, group, SUBG}, resolve this issue by eliminating the effect of $E$.
While solutions based on group balancing are effective when $E$ is observable, they are not feasible when group annotation is not provided. 

Another solution that is effective even in the absence of group annotation is removing the edge $S\xrightarrow{}\tilde{S}$ by intervening on some components in $\tilde{S}$ in order to break the path, as shown in \cref{fig:scm_intervene}.
 More concretely, this solution intervenes on a subset of non-causal components of images without changing the label to reduce their correlation.
 Intervention on $\tilde{S}$ could be done in a more efficient manner if we could set $\tilde{S}$ to a value that has less co-occurrence with $Y$. Such intervention would create a new datapoint that can be assigned to a minority group. Hence, this type of intervention would be a method for upweighting datapoint from minority groups. 

\section{To Which Does ERM Attend More?}
\label{sec:heatmap}
To intervene on the non-causal components of an image, it is essential to determine the causal and non-causal parts of it first. The attribution map of a model trained with standard ERM on an image could be utilized in distinguishing these parts. MaskTune~\cite{Masktune} partly addressed this issue by assuming that for a model trained with standard ERM on a dataset exhibiting spurious correlations, the image parts with high attribution scores are spurious.

However, this assertion does not hold for the majority of realistic datasets. Specifically, the behaviour of a model trained with ERM might vary depending on the easiness of predicting the label from the causal and non-causal parts across different datasets.

Since non-causal parts of images that have spurious correlations with the label, such as the background in the Waterbirds dataset, are shortcuts for models, it is often presumed that a model trained with standard ERM attends more to the non-causal parts of images. However, we show that this assumption does not hold in many real-world scenarios. Across the entire dataset, as opposed to the causal components, non-causal ones do not persistently appear in accordance with the label. Consequently, the causal parts may become generally more predictive. As a result, the easiness of predicting labels from the causal and non-causal parts across the entire dataset determines the focus of ERM.

To illustrate this point, we report the average xGradCAM~\cite{gradcam} score for the foreground and background pixels in the Waterbirds datasets in \cref{fig:gradscore}(a,b) and for the Cifar part and the MNIST part of the Dominoes dataset~\cite{agreedisagree} in \cref{fig:gradscore}(c,d). For the Waterbirds, the average score of the foreground (causal) pixels exceeds that of the background (spurious part) and for the Dominoes, the average score of the spurious part (MNIST part) exceeds that of the causal pixels (i.e. Cifar part). Indeed, if the spurious patterns are easy to learn, the model may attend more to the spurious parts. 
 This tendency becomes more pronounced in samples with low loss, as depicted in \cref{fig:gradscore}.

\section{Method}
Regardless of whether models trained with ERM attend more to the causal or non-causal parts of an image, their attribution map on the image remains useful for distinguishing these two parts. In the following, we introduce a method for identifying the most predictive parts of images using attribution maps.

\subsection{Adaptive Masking by ERM}
\label{sec:adaptive}
Given a model $f_\theta(.)$ that is trained with ERM, and a datapoint $(x, y)$ on which $f$'s loss is low, we want to find a mask that conceals pixels in $x$ except for the most predictive ones. The assumption that the loss on $(x,y)$ is low is necessary since we are more confident that the predictive parts in $x$ are indeed predictive of the correct label. 

\begin{figure}
\raggedright
  \noindent\begin{adjustbox}{max width=0.45\textwidth}
    \begin{tikzpicture}
        \begin{axis}[
        at={(0cm,0cm)},
            xlabel={$p$},
            ylabel={$l_p$},
            y label style={at={(0.07,0.45)}, anchor=south west},
            width=10cm, % Adjust the width as needed
            height=6cm,
            xmin=0, xmax=0.9,
            ymin=0, ymax=0.25, % Adjust ymax according to your data
            xtick={0.1,0.3,0.5,0.7,0.9},
            ytick={0,0.05,0.1,0.15,0.2,0.25}, % Adjust ytick according to your data
            legend pos=north west,
            ymajorgrids=true,
            grid style=dashed,
            axis x line*=bottom,
            axis y line*=left,
            %axis on top,
            clip=false,
            yticklabel style={/pgf/number format/fixed}
        ]
        
        % Provided Y1 data with specified X values
        \addplot[mark=square,blue] coordinates {
            (0, 2.2649509e-05)
            (0.1, 4.6967358e-05)
            (0.15, 4.1722382e-05)
            (0.2, 5.4596363e-05)
            (0.25, 5.5311582e-05)
            (0.3, 7.9747835e-05)
            (0.35, 7.283422e-05)
            (0.4, 0.0001173)
            (0.45, 0.00041715)
            (0.5, 0.0003211)
            (0.55, 0.00025877)
            (0.6, 0.00121627)
            (0.65, 0.00154507)
            (0.7, 0.00341024)
            (0.75, 0.00823874)
            (0.8, 0.01185868)
            (0.85, 0.04502294)
            (0.9, 0.1146403)
        };
        \node[inner sep=0] (sag0) at (axis cs:0.2,0.1) {\includegraphics[width=1.8cm]{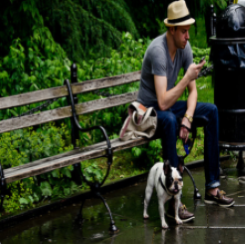}};
        \draw[->] (sag0) -- (axis cs:0,2.2649509e-05);
        
        \node (sag70) at (axis cs:0.6,0.1) {\includegraphics[width=1.8cm]{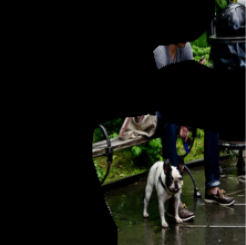}};
        \draw[->] (sag70) -- (axis cs:0.7,0.00341024);

        \node (sag90) at (axis cs:0.8,0.2) {\includegraphics[width=1.8cm]{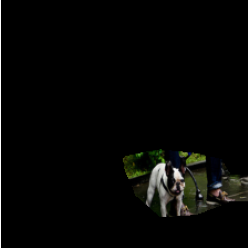}};
        \draw[->] (sag90) -- (axis cs:0.9,0.1146403);
        \end{axis}

        \begin{axis}[
            at={(0cm,5.7cm)},
            xlabel={$p$},
            ylabel={$l_p$},
            y label style={at={(0.07,0.45)}, anchor=south west},
            width=10cm, % Adjust the width as needed
            height=6cm,
            xmin=0, xmax=0.9,
            ymin=0, ymax=0.25, % Adjust ymax according to your data
            xtick={0.1,0.3,0.5,0.7,0.9},
            ytick={0,0.05,0.1,0.15,0.2,0.25}, % Adjust ytick according to your data
            legend pos=north west,
            ymajorgrids=true,
            grid style=dashed,
            axis x line*=bottom,
            axis y line*=left,
            %axis on top,
            clip=false,
            yticklabel style={/pgf/number format/fixed}
        ]

        % Provided Y2 data with specified X2 values
        \addplot[mark=square,blue] coordinates {
            (0, 5.245195e-06)
            (0.1, 0.00029131)
            (0.15, 0.00029131)
            (0.2, 0.00029131)
            (0.25, 0.00029131)
            (0.3, 9.846203e-05)
            (0.35, 0.00417041)
            (0.4, 0.01466554)
            (0.45, 0.02211411)
            (0.5, 0.02081806)
            (0.55, 0.01906955)
            (0.6, 0.02978594)
            (0.65, 0.01721781)
            (0.7, 0.03048482)
            (0.75, 0.21166685)
            (0.8, 0.21166685)
            (0.85, 0.21166685)
            (0.9, 0.21166685)
        };

        \node[inner sep=0] (tru0) at (axis cs:0.2,0.1) {\includegraphics[width=1.35cm]{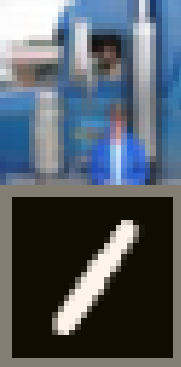}};
        \draw[->] (tru0) -- (axis cs:0,5.245195e-06);
        
        \node (tru70) at (axis cs:0.55,0.13) {\includegraphics[width=1.35cm]{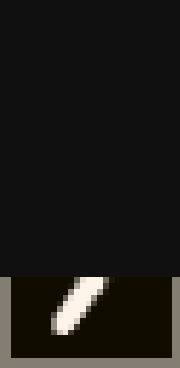}};
        \draw[->] (tru70) -- (axis cs:0.7,0.03048482);

        \node (tru90) at (axis cs:0.83,0.1) {\includegraphics[width=1.35cm]{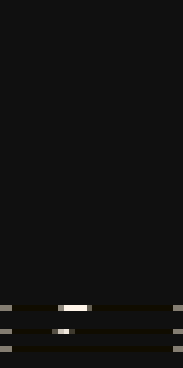}};
        \draw[->] (tru90) -- (axis cs:0.9,0.21166685);
        
        \end{axis}
    \end{tikzpicture}

     \end{adjustbox}

  \caption{Adaptive masking according to the attention scores obtained from the ERM model. The loss value of the masked images for which different portions $p$ of pixels (with the lowest attention score) has been masked is shown as $l_p$. (a) The loss curve for an image of the Dominoes dataset with the label 'truck' on which the ERM model has non-causal attention, and (b) The loss curve for an image of the MetaShift dataset with the label 'dog' on which the ERM model has causal attention.}
  \label{fig:method_adaptive}
\end{figure}
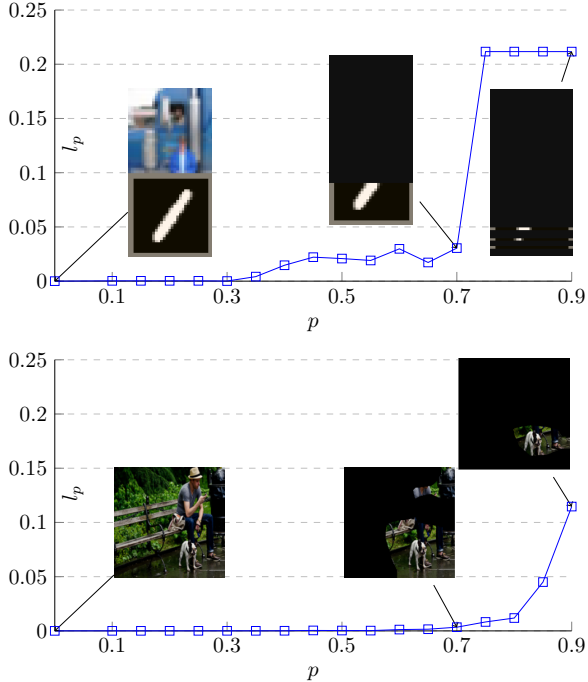

First, the attribution scores of pixels of $x$ are computed using a visual explanation method~\cite{scorecam, gradcam}. Due to its efficiency, we use xGradCAM~\cite{gradcam} for this means and denote the xGradCAM of an input $x$ as $\text {AttributionScore} (x)$.
To have precise adaptive masking, we first define $l(f_\theta(\tilde{x}^p),y)$, in which $l$ is the cross-entropy loss, and $\tilde{x}^p$ is the new image obtained from $x$ by masking out the portion $p$ of pixels with the lowest attribution scores according to $\text {AttributionScore} (x)$. When gradually masking $x$, first the less predictive parts are masked out, which will not have a significant effect on the loss of the masked image. However, when a large proportion of the image, including the predictive part, is masked, the loss increases rapidly, as it is hard for the model to predict the label when the center of attention is (partially) obscured. The effects of gradually masking two images in the Metashift and Dominoes datasets are shown in \cref{fig:method_adaptive}. More precisely, $l(f_\theta(\tilde{x}^p),y)$ can be considered as a function of $p$ whose elbow located at $p^*$ shows the optimal amount of masking for the input $x$. This amount of masking is expected to conceal the non-predictive parts as much as possible while keeping the predictive parts intact. 
Therefore, we define a function $m: \mathbb{R}^{H\times W \times 3} \rightarrow \{0,1\}^{H\times W}$ that returns a mask for its input through an adaptive masking. In fact, $m(x)$ provides a binary mask for $x$ whose value is 0 for a proportion $p^*$ of pixels of $x$ with the lowest attribution score and is 1 for the remaining pixels ($p^*$ denotes the optimal amount of masking found as the elbow of $l(f_\theta(\tilde{x}^p),y)$).

\begin{figure*}[ht!]
\centering
  \noindent\begin{adjustbox}{max width=\textwidth}
  \begin{tikzpicture}[img_block/.style={rectangle, draw, text width=2.1cm, align=center, minimum height=3.4cm},
                      feature_extractor/.style={trapezium, trapezium angle=70, draw, rotate=270, text width=3cm, align=center, minimum height=1cm, fill=gray!30},
                      layer/.style={rectangle, draw, , text width=0.5cm, align=center, minimum height=3.2cm, fill=orange!30},
                      text_box/.style={rectangle, text width=3cm, align=center}]
    
    % Define other blocks

    % Create a bounding box
  \node[draw, inner sep=5pt, fill=purple!20] (box_data) {
    \begin{minipage}{1.5cm} % Adjust the width as needed
      % Include multiple images as nodes in a vertical arrangement
      \includegraphics[width=1.5cm]{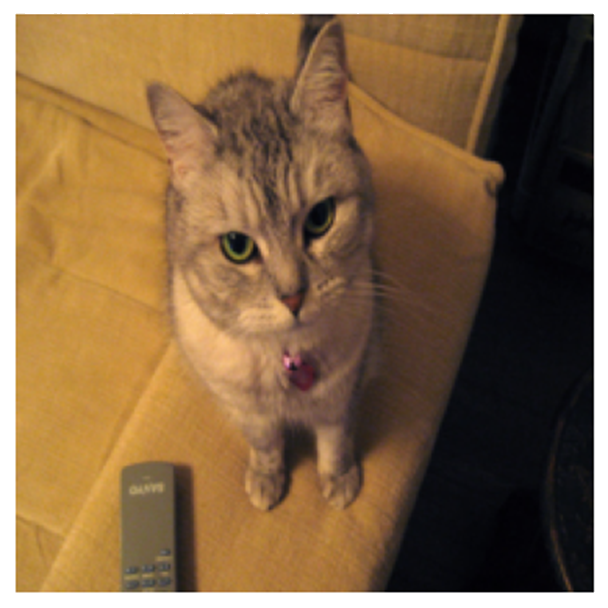}\par
      \vspace{1pt} % Adjust the vertical space between images
      \includegraphics[width=1.5cm]{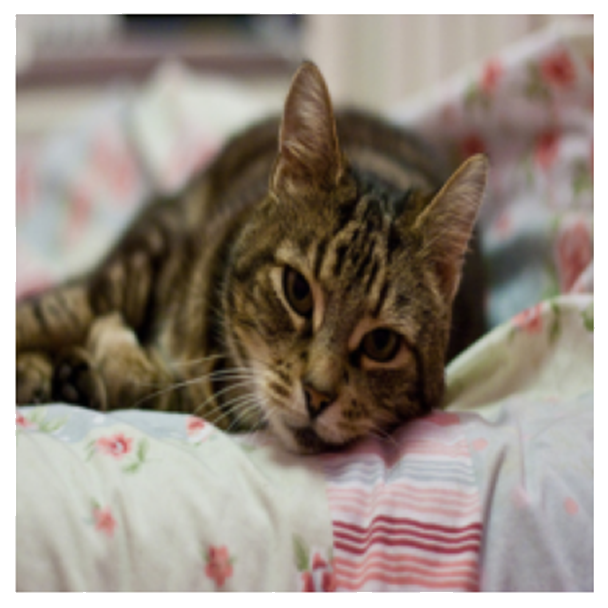}\par
      \vspace{1pt} % Adjust the vertical space between images
      \includegraphics[width=1.5cm]{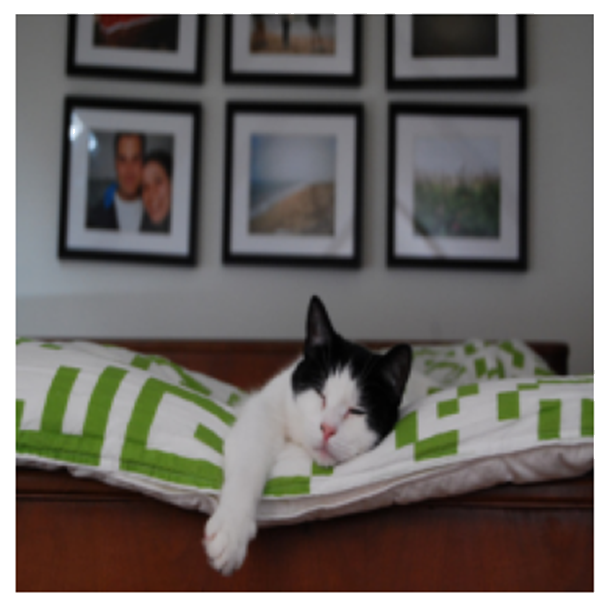}\par
      \hspace{0.7cm}\vdots\par
      \vspace{4pt} % Adjust the vertical space between images
      \includegraphics[width=1.5cm]{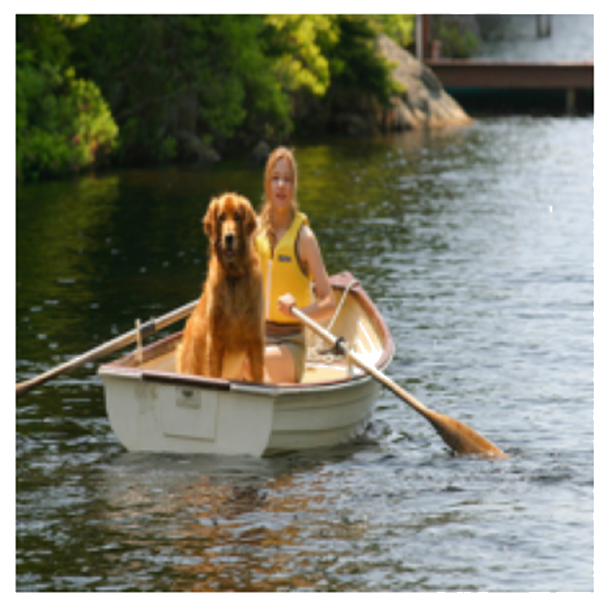}\par
      \vspace{1pt} % Adjust the vertical space between images
      \includegraphics[width=1.5cm]{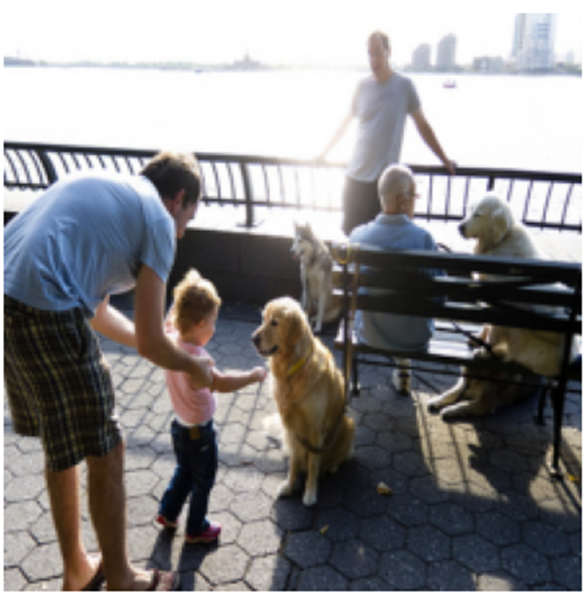}\par
    \end{minipage}
  };

  \node[text_box, text width=0.9cm] (notation5) at ($(box_data.south)-(0,0.3)$) {$\mathcal{B}$};
% Create a bounding box
  \node[draw, inner sep=5pt, right=2cm of box_data, fill=purple!20] (box) {
    \begin{minipage}{1.5cm} % Adjust the width as needed
      % Include multiple images as nodes in a vertical arrangement
      \includegraphics[width=1.5cm]{cat1.png}\par
      \vspace{1pt} % Adjust the vertical space between images
      \includegraphics[width=1.5cm]{cat2.png}\par
      \hspace{0.7cm}\vdots\par
      \vspace{4pt} % Adjust the vertical space between images
      \includegraphics[width=1.5cm]{dog1.png}\par
      \vspace{1pt} % Adjust the vertical space between images
      \includegraphics[width=1.5cm]{dog2.png}\par
    \end{minipage}
  };
  \node[text_box, text width=0.9cm] (notation6) at ($(box.south)-(0,0.3)$) {$\mathcal{B'}$};
    
\node[inner sep=0pt] (img11_input) at ($(box.east)+(1.5,1.5)$) {\includegraphics[width=1.5cm]{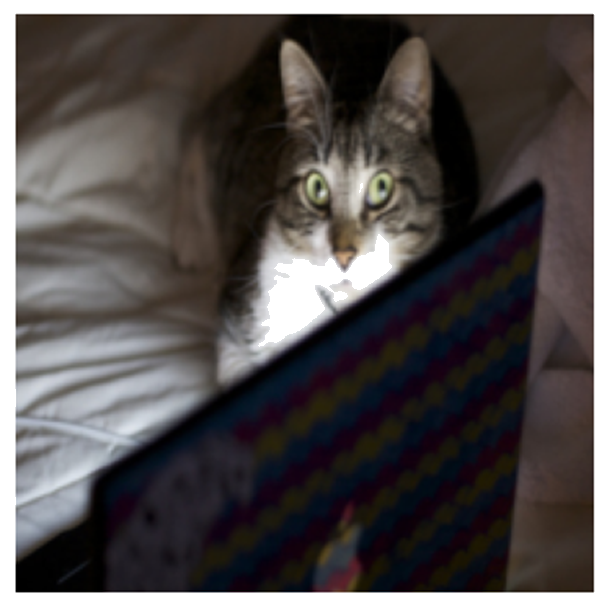}};
\node[below right=-1.7cm of img11_input, inner sep=0pt] (img12_input) {\includegraphics[width=1.5cm]{cat2.png}};
\node[below right=-1.7cm of img12_input, inner sep=0pt] (img13_input) {\includegraphics[width=1.5cm]{cat1.png}};

\node[text_box, text width=0.9cm] (notation7) at ($(img11_input.north)+(0,0.2)$) {$y=0$};

\node[inner sep=0pt, below=1.5cm of img11_input] (img21_input) {\includegraphics[width=1.5cm]{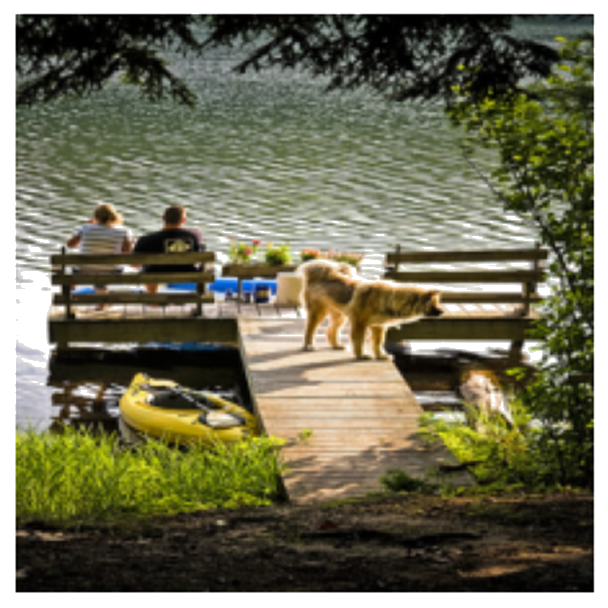}};
\node[below right=-1.7cm of img21_input, inner sep=0pt] (img22_input) {\includegraphics[width=1.5cm]{dog2.png}};
\node[below right=-1.7cm of img22_input, inner sep=0pt] (img23_input) {\includegraphics[width=1.5cm]{dog1.png}};

\node[text_box, text width=0.9cm] (notation8) at ($(img21_input.north)+(0,0.2)$) {$y=1$};

    % Create a bounding box
  \node[draw, dashed] (MaC_box) at ($(img13_input.east)!0.5!(img23_input.east)+(1,0)$)
  {
    \begin{minipage}{1.5cm} % Adjust the width as needed
      Mask and\par
      Combine
    \end{minipage}
  };

    \node[inner sep=0pt] (img31_output) at ($(MaC_box.east)+(2,-0.6)$) {\includegraphics[width=1.5cm]{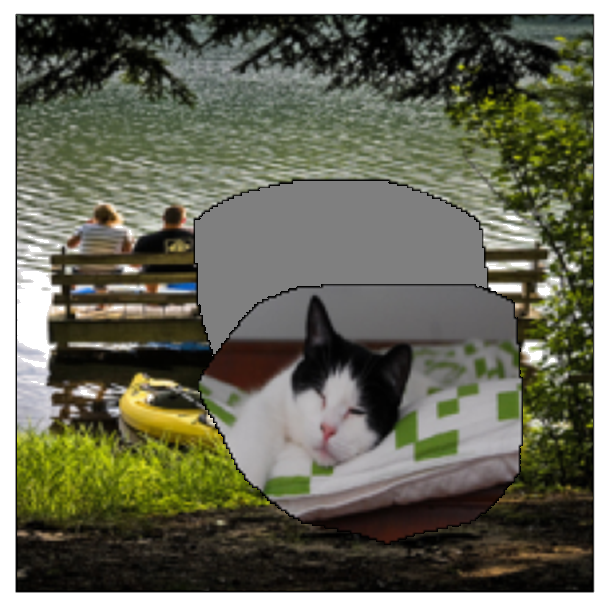}};
\node[above left=-1.7cm of img31_output, inner sep=0pt] (img32_output) {\includegraphics[width=1.5cm]{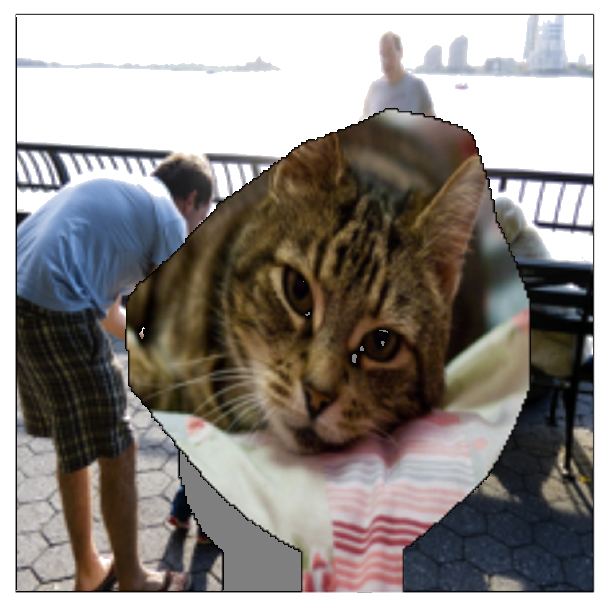}};
\node[above left=-1.7cm of img32_output, inner sep=0pt] (img33_output) {\includegraphics[width=1.5cm]{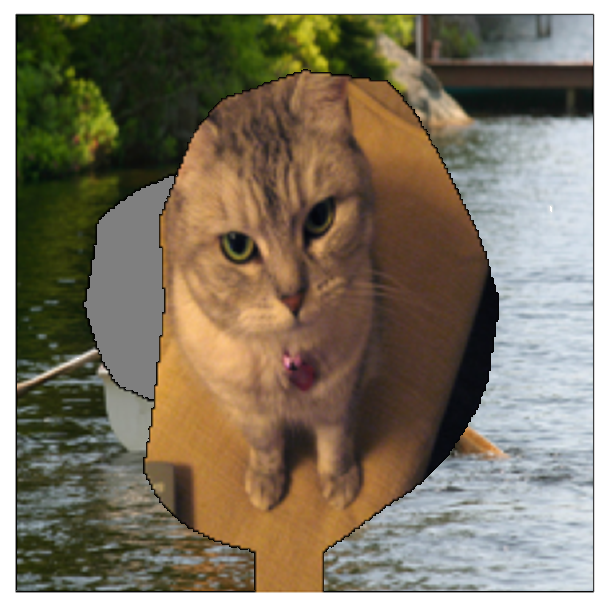}};

\node[text_box, text width=0.9cm] (notation9) at ($(img32_output.south)-(0,0.5)$) {$\mathcal{M}$};

\node[inner sep=0pt, above=1.5cm of img31_output] (img11_output) {\includegraphics[width=1.5cm]{cat3.png}};
\node[above left=-1.7cm of img11_output, inner sep=0pt] (img12_output) {\includegraphics[width=1.5cm]{cat2.png}};
\node[above left=-1.7cm of img12_output, inner sep=0pt] (img13_output) {\includegraphics[width=1.5cm]{cat1.png}};
\node[text_box, text width=0.9cm] (notation11) at ($(img12_output.south)-(0,0.5)$) {$\mathcal{B}$};
\begin{pgfonlayer}{background}
\node[fit=(img31_output) (img13_output), draw, inner sep=10pt, fill=violet!20] (rectangle) {};
\end{pgfonlayer}

    \node[feature_extractor, above right=2.2cm of img31_output.east] (erm_block2) {\rotatebox{90}{ERM}};
    \node[layer, right=0pt of erm_block2.north] (layer_block) at (erm_block2.north) {\rotatebox{270}{Prediction Layer}} ;

    \node[text_box,  text width=0.5cm, align=left] (L_ce) at ($(layer_block.east)+(1.3,0.5)$) {$L_{CE}$};
    \node[text_box,  text width=0.5cm, align=left] (L_comb) at ($(layer_block.east)+(1.3,-0.5)$) {$L_{comb}$};

    \draw[dashed] (layer_block.north)+(2.4,-6.5) -- ++(2.4,3.7);

    \node[rectangle, text width=4cm, align=center] (b_label) at ($(layer_block.north) + (4.3,3.4)$) {b) Mask and Combine};

    \node[rectangle, text width=4cm, align=center] (a_label) at ($(b_label)-(19,0)$) {a) Method Overview};

    \node[text_box, text width=2cm, below=0.1cm of b_label] (notation1) {$x^{(i)},\hspace{1pt}y^{(i)}$};
    \node[draw, below=0.05cm of notation1] (img1_input) {\includegraphics[width=1.5cm]{cat1.png}};
    
    \node[trapezium, trapezium angle=70, draw, text width=0.6cm, align=center, rotate=180, minimum height=0.7cm, fill=gray!30, below=1cm of img1_input] (erm_block12) {\rotatebox{180}{ERM}};

    \node[text_box, below=1cm of erm_block12, fill=pink!30, text width=2.9cm, draw] (ad_box1) {Adaptive Masking};
    \node[text_box, fill=pink!30, text width=2.9cm, right=0.3cm of ad_box1, draw] (ad_box2) {Adaptive Masking};

    \node[trapezium, trapezium angle=70, draw, text width=0.6cm, align=center, rotate=180, minimum height=0.7cm, fill=gray!30, above=of ad_box2] (erm_block22) {\rotatebox{180}{ERM}};
     
    \node[draw, right=of img1_input, above= of erm_block22] (img2_input) {\includegraphics[width=1.5cm]{dog1.png}};
    \node[text_box, text width=2cm, right =of notation1, above=0.05cm of img2_input] (notation2) {$x^{(j)},\hspace{1pt}y^{(j)}$};

    \node[draw, below=0.3cm of ad_box1] (img1_mask) {\includegraphics[width=1.5cm]{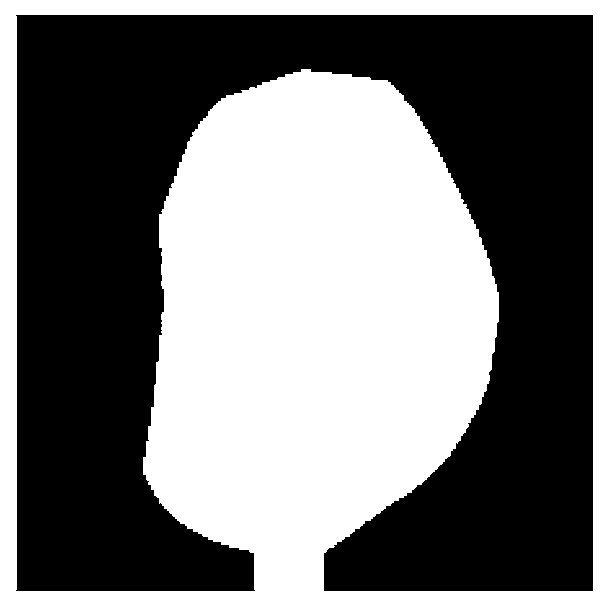}};;
    \node[text_box] (notation3) at ($(img1_mask.south)-(0,0.25)$) {$m(x^{(i)})$};
    \node[draw, below=0.3cm of ad_box2] (img2_mask) {\includegraphics[width=1.5cm]{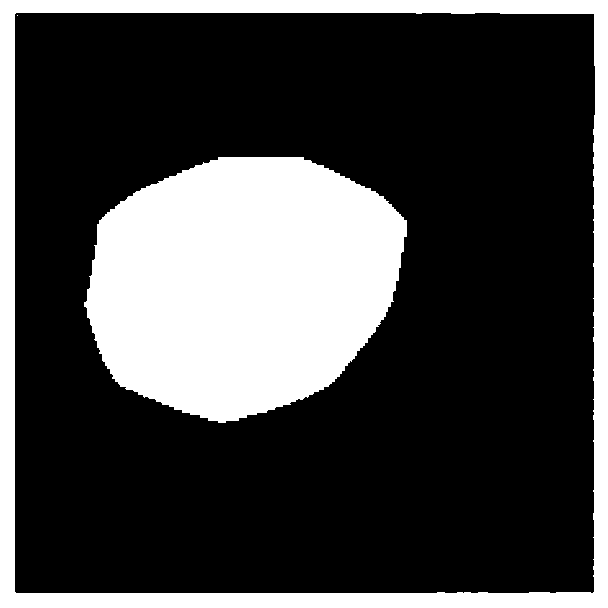}};;
    \node[text_box] (notation300) at ($(img2_mask.south)-(0,0.25)$) {$m(x^{(j)})$};

    \node[rectangle, align=center, text width=1.8cm, fill=yellow!30, draw] at ($(img1_mask.south)!0.5!(img2_mask.south) - (0,0.8)$) (co_box) {\cref{eq:combine}};

    %\node[rectangle, align=center, text width=0.5cm, fill=gray!50, draw, left=of %co_box]  (b_box) {$b$};

    \node[draw, right=of co_box, below=0.2cm of co_box] (img_combine) {\includegraphics[width=1.5cm]{combine1.png}};;
    \node[text_box] (notation4) at ($(img_combine.south)-(0,0.3)$) {$\widehat{x}_{comb_j}^{(i)},y^{(i)}$};

    % Connect blocks with arrows
    \draw[->] (box_data) -- (box) node[above, midway] {Select by};
    \draw[->] (box_data) -- (box) node[below, midway] {Loss Value};
    
    \draw[->] (box_data.north) -- ($(box_data.north)+(0,0.1)$) -| (img13_output.north);
    \draw[->] (box) -- (img11_input);
    \draw[->] (box) -- (img21_input);
    \draw[->] (img13_input) -- (MaC_box);
    \draw[->] (img23_input) -- (MaC_box);
    \draw[->] (MaC_box) -- (img33_output);

    \draw[->] (img31_output.east) -- (erm_block2); 
    \draw[->] (img11_output.east) -- (erm_block2); 
    \draw[->] (img1_input) -- (erm_block12);
    \draw[->] (img2_input) -- (erm_block22);
    \draw[->] (erm_block12) -- (ad_box1);
    \draw[->] (erm_block22) -- (ad_box2);
    \draw[->] (ad_box1) -- (img1_mask);
    \draw[->] (ad_box2) -- (img2_mask);
    \draw[->] (img1_mask) -- (co_box);
    \draw[->] (img2_mask) -- (co_box);
    %\draw[->] (b_box) -- (co_box);
    \draw[->] (co_box) -- (img_combine);
    \draw[->] (layer_block.east) |- (L_ce);
    \draw[->] (layer_block.east) |- (L_comb);
    \draw[->] (img1_input.west) -- ($(img1_input.west)+(-0.8,0)$) |- (co_box.west);
    \draw[->] (img2_input.east) -- ($(img2_input.east)+(0.8,0)$) |- (co_box.east);
        
  \end{tikzpicture}
    \end{adjustbox}

  \caption{(a) An overview of our DaC method. For each batch, a $q$ portion of samples with the lowest loss is selected. Then images of different labels are combined by the \textit{Mask and Combine} module. The overall loss to update the model's last layer parameters is a weighted sum of the loss on the original batch ($L_{CE}$) and the combined data ($L_{\text{comb}}$). The algorithm for this method is shown in \cref{alg:MaC}. (b) The Mask and Combine module. The two input images $x^{(i)}$ and $x^{(j)}$ are masked by \cref{alg: psudocode adaptive}. Afterwards, The selected part of $x^{(i)}$ and the masked parts of $x^{(j)}$ are combined, and the remaining gaps are filled with the mean value of the batch. The new combined image has the same label as $x^{(i)}$ and is used for training the last layer of the model.}
  \label{fig:method}
\end{figure*}

Based on the observation mentioned above, we propose \textit{Adaptive Masking} algorithm shown in \cref{alg: psudocode adaptive} to find the optimal amount of masking for an image. 

\begin{algorithm}
  
  \caption{Adaptive Masking}
  \label{alg: psudocode adaptive}
  
  \SetAlgoLined
  \KwIn{Model $f_{\theta}$; Image $x$; Label $y$; Loss function $l(.,.)$;}
  \KwOut{Mask $m$;}
  
  \BlankLine
  $\delta=0.2$\\
  $x\_score \leftarrow$ AttributionScore($x$)  \;
  \tcp{Calculate $l_p$ for 
  $p\in \{0,\delta,2\delta,3\delta,...,1\}$}
  \For {$p$ \textbf{in} $\{0,\delta,2\delta,3\delta,...,1\}$%$[0, 1]$
  }{
    $\tilde{x}^p \leftarrow$ MaskWithProportion($x\_score$, $p$)\;
    $l_p \leftarrow l(f_{\theta}(\tilde{x}^p),y)$\;
  }
  \tcp{Find the optimal mask}
  $p^* \leftarrow$ FindElbow($l_p$)\;
  $m \leftarrow$ MaskWithProportion($x\_score$, $p^*$)\;
\Return{$m$}\;
  
\end{algorithm}

\subsection{Decompose-and-Compose (DaC)}
\label{sec:main_method}
Viewing images as combinations of components gives us the upper hand of being able to intervene~\cite{pearl} on them. This could be done by combining components of different images to create novel combinations less seen by the model. To be more specific, in the case of spurious correlation, by combining components of different selected inputs, novel images that are more similar to underrepresented data could be created and used during the training of the model as a means of upweighting the underrepresented groups. Here we propose a method for obtaining new datapoints from minority groups, by combining the ones from the majority groups. 
Given dataset $\mathcal{D}$, 
we consider a pretrained classifier $f_\theta(x) = w \circ g_\phi(x)$  with standard ERM on $\mathcal{D}_{tr}$, in which $g_\phi$ is a feature extractor and $w$ is a linear predictor. 
 Afterwards, inspired by~\cite{LastLayer}, we intend to retrain only $w$ on an augmented variation of $\mathcal{D}_{tr}$ that is prepared by intervening on samples to produce new ones to upweight the underrepresented or minority groups.
 
\textbf{Selecting low-loss examples and decomposing them.} For a given training batch $\mathcal{B}$, we first select the subset $\mathcal{B}' = \{(x^{(i)}, y^{(i)}) \in \mathcal{B}|(x^{(i)}, y^{(i)})$ is among $q$ portion of training samples with the lowest loss\} and $q$ is the hyperparameter denoting the portion of the selected samples. 
For each sample $(x^{(i)}, y^{(i)})\in \mathcal{B}'$, first, by \cref{alg: psudocode adaptive}, the mask $m(x^{(i)})$ will be obtained. Then, the two parts of image $x$ is found as $x^{(i)} \odot m(x^{(i)})$ and $x^{(i)} \odot (1-m(x^{(i)}))$. But still, we do not know which of these two parts consists more of the causal regions.\\
As explained in \cref{sec:heatmap}, based on the predictive power of the causal and non-causal parts of the images, the model may attend more to one of them . We inspect both of these assumptions and find which of them yields better results on the validation data. Therefore, we consider $causalflag$ as a hyperparameter in \cref{alg:MaC}, and when this flag indicates the non-causal assumption, masks are inverted, as shown in Lines 10-13, to obtain more causal regions.

\textbf{Composing the causal and non-causal parts of two images.} Given a sample $(x^{(i)}, y^{(i)})\in \mathcal{B}'$, another sample $(x^{(j)}, y^{(j)}) \in \mathcal{B}'$ is selected randomly such that $y^{(i)}\neq y^{(j)}$. Then, we combine the two images as below:
\begin{equation}
\label{eq:combine}
\begin{aligned}
\hat{x}^{(i)}_{\text{comb}_j} & = m(x^{(i)}) \odot x^{(i)}\\ & + (1-m(x^{(i)})) \odot (1-m(x^{(j)}))x^{(j)} \\
& + (1-m(x^{(i)}))\odot m(x^{(j)}) b,
\end{aligned}
\end{equation}
where $b$ is a $1 \times 1 \times 3$ vector indicating the mean of $\mathcal{B}$ across the color channels. 
This formula constructs the combined image by putting the selected parts of $x^{(i)}$ and masked parts of $x^{(j)}$ that are not located on the selected parts of $x^{(i)}$ together, and filling the remaining parts of the image by the default value $b$. The reason for setting the remaining areas equal to $b$ is
to retain the statistics of the batch as much as possible. Finally, we define $\mathcal{M} = \{(\hat{x}^{(i)}_{\text{comb}_j}, y^{(i)}) | (x^{(i)}, y^{(i)})\in \mathcal{B}'\}$ as the combined samples.

As will be shown in Appendix \ref{sec:stats}, most low-loss datatpoints are the ones from the majority groups. By combining the causal and non-causal parts of two majority datapoints from different labels, we make new datapoints from minority groups, thus group-balance the training data without access to group annotation. More precisely, suppose that the model attends more to the causal parts across the dataset. Consider $(x^{(i)}, y^{(i)})$ and $(x^{(j)}, y^{(j)})$ as two samples from majority groups where $x^{(i)} = \psi(\tilde{c}^{(i)}, \tilde{s}^{(i)})$ and $x^{(j)} = \psi(\tilde{c}^{(j)}, \tilde{s}^{(j)})$ and $y^{(i)}\neq y^{(j)}$. Therefore, $\tilde{s}^{(i)}$ and $\tilde{s}^{(j)}$ are in accordance with $y^{(i)}$ and $y^{(j)}$ respectively. Now, in the combined datapoint $\hat{x}^{(i)}_{\text{comb}_j} = \psi (\tilde{c}^{(i)}, \tilde{s}^{(j)})$, the non-causal part $\tilde{s}^{(j)}$ does not have the spurious value corresponding to $y^{(i)}$, which makes this datapoint from the minority groups. It is worth mentioning that this combination tends to be a minority sample independent of whether $m(x^{(i)}) \odot x^{(i)}$ or $(1-m(x^{(i)})) \odot x^{(i)}$ reflects the cause more. 

By the above intervention that combines pairs of data, we generate data in order to break the spurious correlation between the non-causal parts of images and labels. Finally, the loss function during retraining the last layer of the model is defined as below:
\begin{equation}
    L_{CE} = \frac{1}{|\mathcal{B}|}\sum_{(x,y)\in \mathcal{B}} l(f_\theta(x), y),
\end{equation}
% \begin{equation}
%     L_{\text{rmv}} = \frac{1}{|\mathcal{R}|}\sum_{(x,y)\in \mathcal{R}} l(f_\theta(x), y),
% \end{equation}
\begin{equation}
    L_{\text{comb}} = \frac{1}{|\mathcal{M}|}\sum_{(x,y)\in \mathcal{M}} l(f_\theta(x),y)
\end{equation}
\begin{equation}
    L_{\text{total}} = L_{CE}+ \alpha L_{\text{comb}},
\end{equation}
in which $l$ is the cross-entropy loss and $\alpha$ is the hyperparameter determining the importance of the combinations. An overview of our Decompose and Compose (DaC) method in addition to its algorithm is shown in \cref{fig:method} and \cref{alg:MaC} respectively.

\begin{algorithm}
  \caption{Decompose-and-Compose (DaC)}
  \label{alg:MaC}
  \SetAlgoLined
  \KwIn{
  \small
  Model $f_
  \theta (.)= w \circ g_{\phi} (.)$; Dataset $\mathcal{D}_{tr}$;  Loss function $l(.,.)$; Hyperparameters $\alpha,
  q$, causalflag
  }
  
  \BlankLine
%  $\mathcal{L} \leftarrow$ CrossEntropyLoss\
  \For{epoch=$1,2,\dots K$}{
  \For{batch $\mathcal{B}$ in $\mathcal{D}_{tr}$}{
  $b \leftarrow mean(\mathcal B)$\\
  $\mathcal{B'} \leftarrow q$ \textcolor{black}{\text { portion of samples in } $\mathcal{B}$
  \text{ with the lowest loss}}\\%\{ (x,y) \in \mathcal{B} \: | \: l(f_\theta(x),y)<t \}$\\
 % $\mathcal{R} \leftarrow \{\}$
  $\mathcal{M} \leftarrow \{\}$\\
  
  \For{\text{each image} $(x,y) \in \mathcal{B'}$}{
    Pick $(x',y') \in \mathcal{B'}$ s.t. $y\neq y'$\
    
    ${m} \leftarrow$   AdaptiveMasking$(f,x,y,l)$\

    ${m}' \leftarrow$   AdaptiveMasking$(f,x',y',l)$\

    \If{ casualflag=False}{
    $m \leftarrow 1-{m}$\\
    $m'\leftarrow 1-{m}'$\\
        }
%    $\hat{x}_{rmv} \leftarrow m \odot x + (1-m) b$\;
    $\hat{x}_{comb} = m \odot x $\\$+ (1-m) \odot (1-m')x' + (1-m)\odot m' b$\\
    %Add $\hat{x}_{rmv}$ to $\mathcal{R}$ and $\hat{x}_{comb}$ to $\mathcal{M}$\; 
    $\mathcal{M}\leftarrow \mathcal{M} \cup \{(\hat{x}_{comb},y)\}$
  }
  $L_{CE} \leftarrow \frac{1}{|\mathcal{B}|}\sum_{(x,y)\in \mathcal{B}} l(f_\theta(x), y)$\\
%  $L_{rmv} \leftarrow \frac{1}{|\mathcal{R}|}\sum_{(x,y)\in \mathcal{R}} l(f_\theta(x), y)$\;
  $L_{comb} \leftarrow \frac{1}{|\mathcal{M}|}\sum_{(x,y)\in \mathcal{M}} l(f_\theta(x),y)$\\
  $L_{total} \leftarrow L_{CE}+ %\alpha L_{rmv} +
  \alpha L_{comb}$\
  
  $w \leftarrow $ UpdateWeights($L_{total}$)\
  }
}
\end{algorithm}

\section{Experiments}
\subsection{Compared Methods}
\label{sec:baselines} In this paper, we consider 6 baselines besides ERM. 
\textit{DFR}~\cite{LastLayer} argues that even in the presence of spurious correlations, neural network classifiers still learn the core features. Following this, they show that simple retraining of the last layer with group-balanced data can be sufficient to make the model robust to spurious correlation. We evaluated their method on our ResNet50~\cite{resnet} backbone trained with ERM. \textit{Group DRO}~\cite{group} aims to minimize the worst-case loss across groups with strong regularization.
 \textit{LISA}~\cite{LISA} is a data-augmenting technique that aims to learn invariant predictors By intervening on samples with either the same labels but different domains or the same domains but different labels.
\textit{MaskTune}~\cite{Masktune} argues that in the presence of spurious correlations, ERM models attend more to the spurious parts of images. Therefore, they fine-tune an ERM model for one epoch using a masked version of the training data to force the model to focus on the core parts. \textit{CNC}~\cite{correctncontrast} is a contrastive learning method designed to align representations of samples within the same class that have different spurious attributes, while also distinguishing between samples of dissimilar classes that share similar spurious features. \textit{JTT}~\cite{jtt} first uses a model trained with ERM to detect misclassified samples. These samples are subsequently upweighted when training a new model on the dataset.

\begin{table*}[ht!]
\caption{A comparison of mean and worst group accuracy of several methods, including ours, on four datasets. The Group Info column shows whether each method uses group labels of train/validation data, with \cmark\cmark indicating that group info is used in both the training and validation phases. The mean and std are reported over 3 runs on different seeds. The bold and underlined numbers indicate the best results among all methods, and methods not requiring group annotation, respectively.}
\label{tab:method_comparison}
\centering
\small
\begin{tabular}{lccccccccc}
\toprule
& {Group‌~Info}& \multicolumn{2}{c}{Waterbirds} & \multicolumn{2}{c}{CelebA} & \multicolumn{2}{c}{Metashift} & \multicolumn{2}{c}{Dominoes} \\ 
\cmidrule(lr){3-4} \cmidrule(lr){5-6} \cmidrule(lr){7-8} \cmidrule(lr){9-10}
Method  & train/val &Worst & Average & Worst & Average & Worst & Average & Worst & Average \\ 
%\midrule
%ERM & No &  &  &  &  & & &$72.8 _{\pm 1.6}$ & $88.5 _{\pm 0.3}$\\
\midrule
DFR\textsuperscript{*} & \xmark/\cmark\cmark & $92.3 _{\pm 0.2}$ & $93.3 _{\pm 0.5}$ & $88.3 _{\pm 1.1}$ & $91.3 _{\pm 0.3}$ & $72.8 _{\pm 0.6} $ & $77.5_{\pm 0.6}$ & $\boldsymbol{90 _{\pm 0.4}}$ & $\boldsymbol{92.3 _{\pm 0.2}}$\\
Group DRO & \cmark/\cmark & $91.4 _{\pm 1.1}$ & $93.5 _{\pm 0.3}$ & $88.9 _{\pm 2.3}$ & $92.9 _{\pm 0.2}$ & $66.0 _{\pm 3.8}$ & $73.6 _{\pm 2.1}$ & - & -\\
LISA & \cmark/\cmark & $89.2 _{\pm 0.6}$ & $91.8 _{\pm 0.3}$ & $\boldsymbol{89.3 _{\pm 1.1}}$ & $92.4 _{\pm 0.4}$ & $59.8 _{\pm 2.3}$ & $70.0 _{\pm 0.7}$ & - & -\\
\midrule
MaskTune\textsuperscript{*} & \xmark/\xmark & $86.4 _{\pm 1.9}$ & $93.0 _{\pm 0.7}$ & $79.4$ & $89.5$ & $66.3 _{\pm 6.3}$ & $73.1 _{\pm 2.2}$ & $65.8 _{\pm 4.7}$ & $85.6_{\pm  0.7}$\\
CnC & \xmark/\cmark & $88.5 _{\pm 0.3}$ & $90.9 _{\pm 0.1}$ & $\underline{88.8 _{\pm 0.9}}$ & $89.9 _{\pm 0.5}$ & - & -& -& -\\
JTT & \xmark/\cmark & $86.7$ & $93.3$ & $81.1$ & $88.0$ & $64.6 _{\pm 2.3}$ & $74.4 _{\pm 0.6}$ & - & - \\
\midrule
Base (ERM) & \xmark/\xmark & $70.8_{\pm 0.5}$ & $91.6 _{\pm 0.1}$& $41.7$ & $\underline{\boldsymbol{96.0}}$ & $61.3 _{\pm 3.4}$  & $73.9 _{\pm 1.5}$ & $72.8 _{\pm 1.6}$ & $88.5 _{\pm 0.3}$ \\ 
DaC-C & \xmark/\cmark &$\underline{\boldsymbol{92.6 _{\pm 0.2}}}$ & $94.9_{\pm 0.2}$ & $76.11_{\pm 0}$& $91.35_{\pm 0.2}$& $76.0 _{\pm 0.8}$ & $\underline{\boldsymbol{80.0 _{\pm 1.4}}}$ & $89.0 _{\pm 0.7}$ & $92.2 _{\pm 0.2}$\\
DaC & \xmark/\cmark & $92.3 _{\pm 0.4}$ & $\underline{\boldsymbol{95.3 _{\pm 0.4}}}$ & $81.9 _{\pm 0.7}$& $91.4 _{\pm 1.1}$ & $\underline{\boldsymbol{78.3 _{\pm 1.6}}}$ & $79.3 _{\pm 0.1}$ & $\underline{89.2 _{\pm 0.1}}$ & $\underline{92.2 _{\pm 0.3}}$ \\
\bottomrule
\end{tabular}
\end{table*}

\subsection{Setup}
The experiments are done on four datasets: Waterbirds~\cite{group}, CelebA~\cite{celeba}, Metashift~\cite{metashift}, and Dominoes~\cite{agreedisagree}. The details for these datasets are in Appendix \ref{sec:details}.

\label{sec:setup}
Similar to all the works mentioned in \cref{sec:baselines}, the model we use in our experiments is ResNet-50 pre-trained on ImageNet. For ERM training, on all datasets except Dominoes, we used random crop and random horizontal flip as data augmentation, similar to \cite{Masktune,LastLayer}. Retraining the last layer of the model did not require data augmentation. Also, to reduce the strong disturbance of class imbalance, we used class-balanced data to retrain the last layer on CelebA, which is the same approach we took to reproduce the results of \cite{Masktune}. Model selection and hyper-parameter fine-tuning are done according to the worst group accuracy on the validation set. For all the datasets, the value for $\alpha$ and the proportion $q$ of the selected data (according to their loss) for combining have been chosen from $\{1, 2, \hdots 10\}$, and $\{0.2, 0.4, 0.5, 0.6, 0.8, 1\}$ respectively. For adaptive masking, we used \cite{kneelPython} python implementation to determine the optimal amount of masking.

In addition to the main method (DaC), we test another version of our method, named DaC-C, which uses all the correctly classified samples for making combined data, and removes the hyperparameter $q$. Thus, it uses correct classification as a way for selecting low-loss samples.

For all datasets, we have trained the model in two settings: one by assuming that the model generally attends to the causal parts, and the other by the assumption that the model trained by ERM attends more to the non-causal parts. For all datasets except the Dominoes, the former has better worst group accuracy on the validation set.

The details for training the base ERM model and training the last layer of the model with DaC are in Appendix \ref{sec:details}.

\subsection{Results}
\label{sec:results}
The results of our experiments along with reported results for DFR~\cite{LastLayer}, Masktune~\cite{Masktune}, LISA~\cite{LISA}, Group DRO~\cite{group}, and JTT~\cite{jtt} on four benchmarks are illustrated in Table \cref{tab:method_comparison}. Both the worst group accuracy and the average group accuracy as the most commonly used metrics to evaluate robustness again spurious correlation have been reported. Similar to \cite{LastLayer}, the Group Info column shows whether the label of the group (majority/minority) to which datapoints belong is available for training or validation data. Among the methods in \cref{tab:method_comparison}, only DFR requires group info of validation data during the training phase (and not just for model selection), which is shown by \cmark\cmark.\\
The results of the methods annotated with $*$ are reproduced by our own experiments.
As for the other methods, the results on the Waterbirds and CelebA datasets are from their original paper. The results for Metashift are reported by ~\cite{discoverandcure}. %figures
 Three methods among the baselines, i.e. DFR, Group DRO, and LISA need the group label during the training phase as mentioned in \cref{tab:method_comparison}. According to these results, our method outperforms other methods that don't require group labels during training with a large margin in both mean and worst group accuracy metrics on Waterbirds, Dominoes, and Metahift datasets. Moreover, although the proposed method does not need the group label of the training data, it outperforms Group DRO and LISA on Waterbirds and Metashift datasets and is on par with DFR on these datasets.
It is worth mentioning that the CelebA dataset does not match the type of spurious correlation for which our method has been designed as mentioned in \cref{sec:preliminaries}. More precisely, in addition to the face (including gender features) that can be considered as a non-causal part for classifying the hair colour, there is also a spurious attribute for the causal part (i.e. hair) in this problem. There is a spurious correlation between the volume of the hair and the label, as will be discussed in Appendix \ref{sec:details}.

Unlike most previous methods, which usually do not generalize well to samples with diversity shift, our method can perform well also for diversity shift if it is due to novel compositon in the scene. For more analysis of DaC-C, please refer to Sec. \ref{sec:corr}.

\subsection{Ablation Study}
\subsubsection{Effect of Combining Images}
\begin{figure}[h!]
\centering
	\begin{tikzpicture}
		\begin{axis}[
			xlabel={$\alpha$},
			ylabel={Worst group accuracy},
			legend style={at={(1,0.31)}, legend columns=2,  font=\footnotesize},
			grid=major,
			width=8 cm, height=4.6 cm, every axis plot/.append style={ultra thick}
			] 
							\addplot[ color=brown] coordinates {
				(0, 81.57 )
				(1, 89.98)
				(2, 91.78)
				(3, 92.18)
				(4, 92.99)
				(5, 93.39)
				(6, 93.39)
				(7, 93.39)
				(8, 93.79)
				(9, 93.4)
				(10, 93.4)
			};
			\addlegendentry{Dominoes}
			\addplot[ color=OliveGreen] coordinates {
				(0, 70.7 )
				(1, 85.7)
				(2, 86.5)
				(3, 86.5)
				(4, 86.5)
				(5, 86.5)
				(6, 86.5)
				(7, 86.5)
				(8, 86.5)
				(9, 86.5)
				(10, 87.2)
			};
			\addlegendentry{Waterbirds}
			\addplot[color=RedViolet] coordinates {
        (0, 58 )
(1, 76.3)
(2, 76)
(3, 77.3)
(4, 77.3)
(5, 76.6)
(6, 78.3)
(7, 78.6)
(8, 75.6)
(9, 72)
(10, 79)
			};
			\addlegendentry{Metashift}
			\addplot[color=MidnightBlue] coordinates {
        (0, 84.62 )
(1, 84.57)
(2, 82.42)
(3, 82.97)
(4, 83.52)
(5, 84.07)
(6, 83.89)
(7, 83.49)
(8, 83.05)
(9, 82.67)
(10, 82.11)
			};
			\addlegendentry{CelebA}
		\end{axis}
	\end{tikzpicture}
    \caption{Worst group accuracy on different datasets with respect to $\alpha$. $\alpha\geq 1$ is enough to increase worst group accuracy rapidly.}
    \label{fig:alpha_Dominoes}
\end{figure}
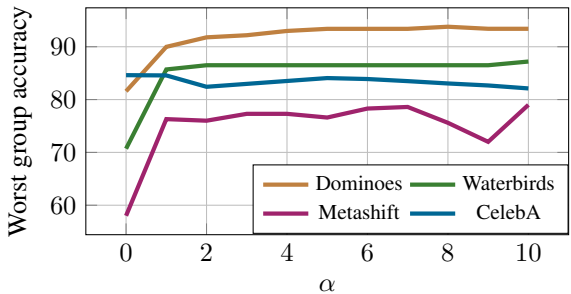

Combining images proves to be extremely effective in enhancing the worst test group accuracy, which is evident in \cref{fig:alpha_Dominoes}. It can be seen that the value of $\alpha\geq1$ highly reduces the reliance on spurious patterns and also the range of the proper value of $\alpha$ is similar in different datasets.

\subsubsection{Effect of the Proportion of the Selected Data}
\label{sec:corr}
As the number of the selected samples for combining increases, more datapoint on which the base model has a higher loss will be used for combining. The model is more prone to attending to irrelevant parts in these samples, which results in the combined images being wrongly labelled. As shown in \cref{tab:method_comparison}, the model has the best worst group accuracy when we choose the proportion of the selected samples by hyper-parameter tuning. Selecting all the correctly-classified samples omit the need for tuning hyperparameter $q$,  but it may degrade the accuracy of the method when the ERM overfits on the training samples, since in this case, all samples may be used for combination.

\section{Conclusion}
In this paper, we first showed that whether the models trained with standard ERM pay more attention to the causal or spurious parts of images in a dataset depends on the predictiveness of these parts across the entire dataset. Furthermore, in most realistic datasets, due to the lower correlation of non-causal parts with the label compared to the causal ones, ERM usually shows causal attention.
We then utilized attribution maps of an ERM model on images to decompose them and find the significantly more attended parts, by monitoring the classification loss of the ERM model on masked images. According to this decomposition of images, we also suggested a method for combining images with low loss which helps to mitigate the spurious correlation and diversity shift. This method has proven to be highly effective on four benchmarks and has a comparable performance with methods that require minority/majority group annotation of training data, unlike ours.
Although this research was primarily focused on spurious correlations between parts of images and labels, the idea could potentially be extended to more complex scenarios where there is a spurious correlation between attributes of the objects in a scene and the label. Further research on more accurate methods for distinguishing causal and non-causal parts, and more advanced interventions on images is left for future work.

\section*{Acknowledgments}
We would like to thank Mr. Mohammad-Mahdi Samiei for his insightful and constructive comments.
\clearpage
{
    \small
    \bibliographystyle{ieeenat_fullname}
    \bibliography{main}
}
\clearpage
\setcounter{page}{1}
\maketitlesupplementary

\section{Related Work}
\subsection{Mitigating Spurious Correlation}
It has long been known that deep models trained under standard ERM settings are vulnerable to spurious correlations~\cite{bias, group, irm}. This problem has been addressed in the literature under terms such as shortcut learning~\cite{shortcutlearning1, shortcutlearning2} and simplicity bias~\cite{simplicitybias1, simplicitybias4}. \\
Most well-known works in the literature approach mitigating spurious correlation by either group balancing or sample reweighting. Group DRO~\cite{group}, which is one of the best-performing methods proposed so far, uses group annotations to minimize the worst group error. SUBG~\cite{SUBG} trains a model by ERM on a random group-balanced subset of data and has proven to be effective on several benchmarks. Following~\cite{SUBG}, 
DFR~\cite{LastLayer} states that models trained with ERM are capable of extracting both core and non-core features of an image and proposes to retrain only the last layer of the predictor on a group-balanced subset of training or validation set to make models robust to spurious correlation. Although these methods have acceptable performance, they require group labels of the training or validation set for the training. This assumption is not feasible in many scenarios and has been addressed by several methods that aim to train robust models without access to the group labels. Among these methods, some introduce methods for reweighting or pseudo-labelling samples for last layer retraining~\cite{AFR, disagreement}. AFR\cite{AFR} upweights samples for which a model trained with ERM assigns a low probability to the correct class. DD-DFR~\cite{disagreement} assigns pseudo group labels to samples based on the change of the model's prediction on them when adding dropout to the model. In addition to this line of work, \cite{jtt,LFF} introduce methods for fine-tuning whole models without group knowledge by upweighting or upsampling the data misclassified by a model trained by ERM. JTT~\cite{jtt} upsamples datapoints which are misclassified by a model trained by ERM, with the assumption that these samples are mostly from under-represented groups.\\\\
The methods mentioned above have a common assumption that the misclassified samples or samples with high loss are mostly from minority groups. While this holds true for some samples, in many cases, the main reason behind the model's high loss on a sample is the complexity of its causal regions. On the contrary, we manually make combined images that are theoretically from the minority groups, as a means for upweighting under-represented samples.

\subsection{Data Augmentation for Bias Mitigation} A line of work uses data augmentation for enhancing models' generalizationability~\cite{Advmixup,  LISA, discoverandcure}. Inspired by mixup~\cite{mixup}, LISA~\cite{LISA} selectively interpolates datapoints across different groups or different labels to train an invariant predictor. DISC~\cite{discoverandcure} utilizes a concept bank to detect spurious concepts and intervenes on samples using these concepts to balance the spurious attributes. In addition to these works, few works use synthetic data augmentation for balancing the training data~\cite{USB, ASB, biswap, ffr}. GAN debiasing~\cite{ASB} uses a GAN to generate images and intervene on them in the latent space. FFR~\cite{ffr} combines synthetic data augmentation and loss-based debiasing methods (such as Group DRO~\cite{group}) for mitigating spurious correlation.\\\\
Almost all the methods based on data augmentation require the knowledge of the spurious attributes or group labels, or use additional concept banks or generative models for detecting and intervening on spurious attributes. DaC on the other hand, augments the training data with none of the mentioned requirements.

\subsection{Attention-based Masking for Out-of-Distribution Generalization}
Some other works were proposed for removing the irrelevant parts of images by masking ~\cite{mask1, mask2}. CaaM~\cite{caam} proposes a causal attention module that generates data partitions and removes confounders progressively to enhance models' generalizability. \cite{maskfinetune} masks patches of images based on the class activation map and refills them from patches of other images and utilizes these samples for representation distillation with a pretrained model. 
 Decoupled-Mixup~\cite{decoupled} distinguishes discriminative and noise-prone parts of images and fuses these parts by mixup separately. MaskTune~\cite{Masktune} which is the most similar work to ours, based on the assumption that models trained with ERM mostly focus on parts of the image with high spurious correlation to the label, masks parts of the image with the highest scores according to xGradCAM. Then a new model is fine-tuned on the masked data.\\\\
None of the methods mentioned above, except MaskTune, strive to extract the causa parts of images in order to determine the true label of the newly obtained images. However, a key point in DaC is that it distinguishes the causal parts from the non-causal regions to be able to make combined images and determine their label. Additionally, as discussed in \cref{sec:heatmap}, it cannot be simply assumed that the focus of models trained with ERM is on non-causal parts of images, which is the most noticeable downfall of MaskTune, that we aimed to solve to an extent.

\section{Details on Experiments}
\label{sec:details}
\subsection{Datasets}
In this study, We compared methods on four datasets with distribution shifts. The first three datasets are related to correlation shift and the last one includes diversity shift between the train and test sets according to the categorization introduced in \cite{oodbench}.\\
\textbf{Waterbirds} This dataset is created by combining bird photos from the Caltech-UCSD Birds-200-2011~\cite{cub} dataset with image backgrounds from the Places dataset~\cite{places}. The birds are labelled as either waterbirds or landbirds and are placed on either water or land backgrounds. Waterbirds are more frequently shown on water backgrounds, while landbirds are more often shown on land~\cite{group}.
\\
\textbf{CelebA} CelebA celebrity face dataset in the presence of spurious correlations was proposed by \cite{group}. In this dataset the binary label is assigned to the hair colour and the gender is the attribute with spurious correlation with the label~\cite{celeba}.
\\
\textbf{Dominoes}: This dataset, synthesized in a manner similar to \cite{agreedisagree}, consists of paired images: one from CIFAR10 and one from MNIST. The CIFAR10 image, either an automobile or a truck, serves as the target label. Meanwhile, the MNIST image, a zero or a one, acts as the spurious part. The spurious correlation between MNIST digits and the label is 90\%.
\\
\textbf{Metashift}: Our setup for Metashift dataset follows ~\cite{discoverandcure}. The target is to classify cats and dogs, and spurious features are objects and backgrounds, namely sofa, bed, bench, and bike. The test images are from backgrounds that are not present in the training set. 

\subsection{Details on the CelebA Dataset}
\label{sec:celeba}
\begin{figure}
\centering
\includegraphics [width = 0.4 \textwidth]{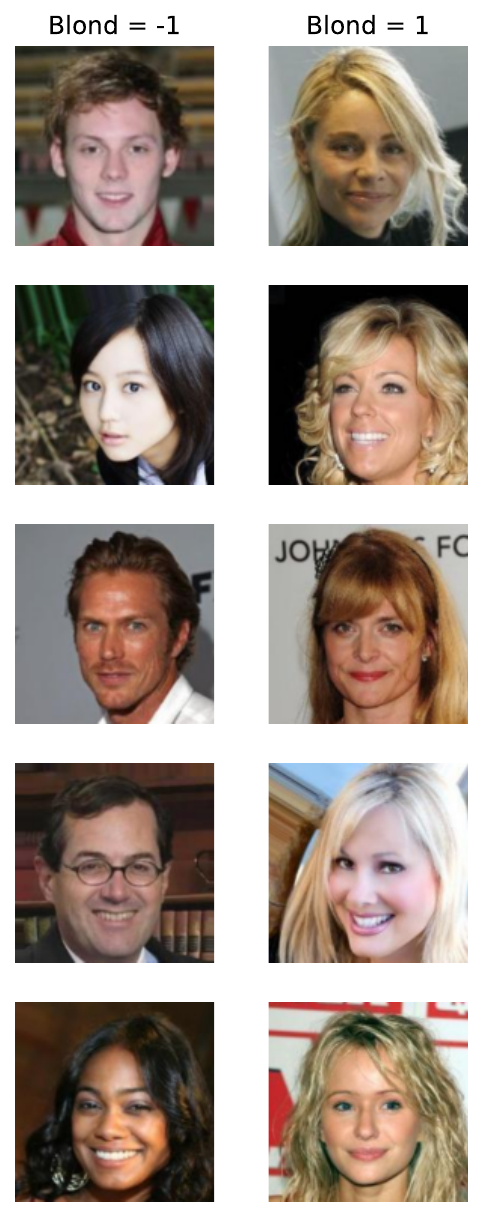}
\caption{Some samples from the CelebA dataset.}
\label{fig:celeba_hair}
\end{figure}

As mentioned in \cref{sec:results}, in addition to the spurious correlation between gender (which can be inferred from the facial features) and hair colour, some hair attributes contribute to hair volume such as hair wave and baldness, which are correlated with the hair colour. The number of people with each hair colour and specific attributes is extracted from the CelebA metadata and shown in \cref{tab:wavy,tab:bald}. According to the statistics, while about $0.05\%$ of blond people wear hats or are bald, more than $8$ per cent of people who are not blond wear hats or are bald. Similarly, the percentage of blond people with wavy hair is more than $1.5$ times greater than the ones that are not blond. Additionally, our eye observations from the dataset indicate that there is a correlation between the length of hair and its colour, as short hair is more co-occurred with non-blond hair. It is worth mentioning that since the attribute of hair length was not available in the CelebA metadata, we assessed this claim by eye observation. A few examples of randomly selected samples from each hair colour are shown in \cref{fig:celeba_hair}.

\begin{table}[h!]
\caption{Number of people with wavy hair with each hair colour.}
\label{tab:wavy}
\centering
    \begin{tabular}{c|cc}
     & Blond = -1 & Blond = 1\\
    \hline
    Wavy = -1 & 121761 & 16094\\
    Wavy = 1 & 50855 & 13889 \\
    \end{tabular}
\end{table}

\begin{table}[h!]
\caption{Number of people with each hair colour that are bald or wear a hat.}
\label{tab:bald}
\centering
    \begin{tabular}{c|cc}
     & Blond = -1 & Blond = 1\\
    \hline
    Bald = -1 $\land$ Hat = -1 & 158440 & 29817\\
    Bald = 1 $\lor$ Hat = 1 & 14176 & 166 \\
    \end{tabular}
\end{table}

\subsection{ERM Training Details}
Similar to \cite{LastLayer}, we used SGD optimizer with learning rate $10^{-3}$ and momentum $0.9$ for all datasets. We used weight decay $10^{-3}$ for Waterbirds, Metashift and Dominoes dataset and $10^{-4}$ for CelebA. The batch size for CelebA, Waterbirds, Metashift, and Dominoes were $128$, $32$, $16$, and $16$ respectively. The model was trained for $100$ epochs on the Waterbirds and Metashift datasets, and for $30$ and $15$ epochs on the CelebA and Dominoes. 

\begin{table}[h!]
\caption{Hyperparameters for DaC}
\label{tab:dac_details}
\centering
    \begin{tabular}{lccc}\toprule
    \multirow{ 2}{*}{Dataset} & \multicolumn{3}{c}{Hyperparameters}
    \\\cmidrule(lr){2-4} & epochs & $\alpha$ & $q$\\\midrule
    Waterbirds  & 20 & 10 & 0.6
    \\CelebA  & 15 & 5 & 0.2
    \\MetaShift  & 30 & 6 & 0.5 
    \\Dominoes  & 20 & 6 & 0.8 
    \\\midrule
    \end{tabular}
\end{table}

\begin{table*}[ht!]
\caption{Mean and worst group accuracy on the validation sets of four datasets when applying DaC using the original or inverted masks.}
\label{tab:inv_mask}
\centering
\begin{tabular}{cccccccccc}
\toprule
Invert & \multicolumn{2}{c}{Waterbirds} & \multicolumn{2}{c}{CelebA} & \multicolumn{2}{c}{Metashift} & \multicolumn{2}{c}{Dominoes} \\ 
\cmidrule(lr){2-3} \cmidrule(lr){4-5} \cmidrule(lr){6-7} \cmidrule(lr){8-9}
    Mask &Worst & Average & Worst & Average & Worst & Average & Worst & Average  \\\midrule
   	\xmark 
		& $88.4$ & $93.1$ & $84.6$& $91.2$ & $79$ & $79$ & $19.6$& $63.1$\\
		\cmark & $22.6$& $63.2$& $83.9$& $90.1$ &$45$ &$60.1$  & $89.2$ & $93.0$\\
\bottomrule
\end{tabular}
\end{table*}

\subsection{DaC Training Details}
For all datasets, Adam optimizer with a learning rate of $0.5\times10^{-2}$, and step learning rate scheduler with step size $5$ and gamma $0.5$ were used. The batch size was 64 for all datasets. To encourage the diversity of training data during retraining the last layer, in cases when the selected samples with low loss in each batch were only from one class, we randomly combined the selected images with others from the same class. No regularization terms were used for retraining the last layer of the model. The proportions for creating the curve of the loss with respect to the amount of masking in adaptive masking did not contain 1, since masking the whole image would trivially increase the loss on the masked image significantly. More details regarding the number of epochs, and optimal values for $\alpha$ and $q$ are in \cref{tab:dac_details}. Batch size, $\alpha$ and $q$ were selected from $\{32, 64\}$, $\{1,...,10\}$, and $\{0.2, 0.4, 0.5, 0.6, 0.8, 1\}$ respectively, and the criteria for hyperparameter selection was the worst group accuracy on the validation set.

\subsection{Original Masks or Inverted Ones?}
\label{sec:reverse_results}
As mentioned in \cref{sec:main_method}, we train the model in two settings corresponding to the ERM casual attention and ERM non-causal attention assumptions. For the former setting, i.e. ERM casual attention, we keep the parts obtained by adaptive masking as the causal parts while for the later one, i.e. ERM non-causal attention, the parts remained by adaptive masking are considered as non-causal and thus we invert the masks in order to obtain the causal regions for DaC. Based on the worst group accuracy of the model trained by DaC on the validation set in these two settings, it can be determined whether the parts to which the model generally pays more attention are causal or non-causal. The results for both cases are in \cref{tab:inv_mask}. According to the results, unlike the Dominoes, on Waterbirds and Metashift the model attends more to the causal components. Regarding the CelebA dataset, it seems that the attention of the model does not grasp the entire hair parts in the image, hence, the inverted mask still contains a proportion of the hair. This was also reflected in the results in \cref{tab:method_comparison}, in which, unlike other datasets, our model has a lower performance on CelebA. For more details on the CelebA dataset, refer to \cref{sec:celeba}.

\subsection{Details on the Kneedle Algorithm}
As mentioned in \cref{sec:setup}, we use the \textit{Kneedle} algorithm for finding the optimal amount of masking in \cref{alg: psudocode adaptive}. This optimal amount is indicated by the \textit{elbow} (i.e. the point with the highest curvature) of the curve of the loss with respect to the amount of masking. Since we only have access to a finite number of points from this curve, we use the Kneedle algorithm, which identifies elbows in a finite set of points from a curve. 

The Kneedle method is based on the concept that knee points approximate the local maxima when the set of points is rotated about a specific line. This line is determined by the first and last points and is chosen to preserve the overall behaviour of the set. By rotating the curve about this line, knee/elbow points are identified as the points where the curve deviates most from the straight line segment connecting the set's endpoints. This approximation effectively captures the points of maximum curvature for the discrete set of points. The algorithm works as follows:

\begin{enumerate}
    \item Smoothing: it applies a smoothing spline or other smoothing methods to data.
    \item Normalization: It normalizes smoothed data by min-max normalization to function well regardless of the magnitude of data values.
    \item Difference Computation: It defines 
    $D_d$ as the set of differences between x- and y- values. The knee is where the difference curve changes from horizontal to sharply decreasing.
    \item Local Maxima Calculation: It identifies the local maxima of the difference curve as candidate knee points.
    \item Threshold Calculation: For each local maximum $(x_{lmx_i}, y_{lmx_i})$ in the difference curve it defines the 
    $T_{lmx_i}$ which is based on the average difference between consecutive $x$ values in the difference curve and a sensitivity parameter, $S$. This parameter determines how aggressive the method is. Smaller values for $S$, detect knees quicker, and large values are more conservative.

    \begin{center}
        $T_{lmx_i} = y_{lmx_i} - S. \frac{\sum_{i=1}^{n-1} x_{i+1} - x_i}{n-1}$
    \end{center}

    \item Knee Declaration: If any difference value $(x_i, y_i)$, where $j > i$, drops below the threshold $y = T_{lmx_i}$ before the next local maximum in the difference curve is reached, the method declares that local maximum as a knee point.

    Kneedle's run time for any given n pairs of $x-$ and $y-$ values is bounded by $\mathcal{O}(n^2)$.
\end{enumerate}

\subsection{Training Time}
Since the ERM model used for computing the attribution scores of the pixels is fixed, extracting the attention heatmap and adaptive masking is done as a preprocess. Hence, during training, the previously prepared and saved masks are used, similar to MaskTune~\cite{Masktune}.
 Additionally, since the optimal percentage of the masked pixels in \textit{Adaptive Masking} is selected among a small number of candidates, the time complexity of FindElbow is constant. The training time of several methods (excluding the ERM phase of the methods) on Waterbirds is shown in \cref{tab:time}.
 
\begin{table}[h!]
\centering
\caption{The training time of different methods (excluding the ERM training phase) on the Waterbirds dataset on Nvidia A100 GPU}
\label{tab:time}
\begin{tabular}{l|ccccc}
 Method & DFR & CnC & JTT & MaskTune & Ours \\ 
 \midrule
Time \footnotesize{(min)}         &  4   &  85   &   58  &   6.5   & 18.9      \\ 
\end{tabular}
\end{table}

\section{More Empirical Observations}
\label{sec:stats}
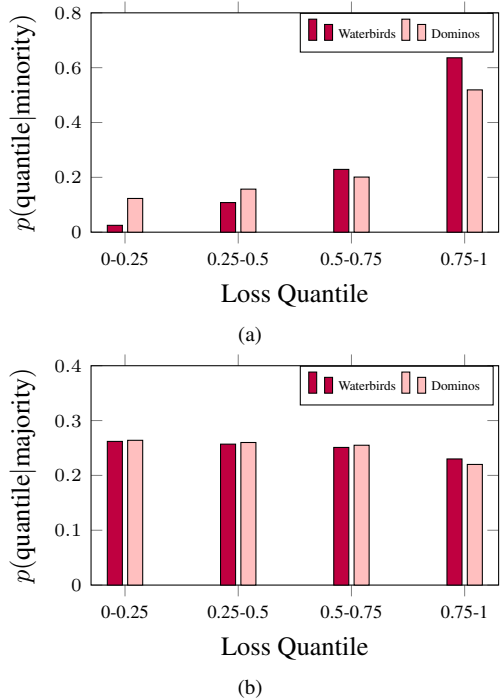
\begin{figure}[h!]
    \centering
    \begin{subfigure}{0.4\textwidth}
        \centering
        \begin{tikzpicture}
        \begin{axis}[
            ybar,
            bar width=0.2cm,
            width=7cm,
            height=4.5cm,
            legend style={at={(0.75,1)}, anchor=north, legend columns=-1, font=\tiny},
            symbolic x coords={0-0.25, 0.25-0.5, 0.5-0.75, 0.75-1},
            xtick=data,
            xlabel={Loss Quantile},
            ylabel={$p(\text{quantile}|\text{minority})$},
            ymin=0,
            ymax=0.8,
            nodes near coords={},
            tick label style={font=\fontsize{7}{9}\selectfont}
        ]

            \addlegendentry{Waterbirds}
            \addlegendentry{Dominos}

            % Waterbirds
            \addplot[fill=purple] coordinates {(0-0.25, 0.025) (0.25-0.5, 0.108) (0.5-0.75, 0.229)  (0.75-1, 0.636)};
            
            % Dominos 
            \addplot[fill=pink] coordinates {(0-0.25, 0.123) (0.25-0.5, 0.157) (0.5-0.75, 0.201) (0.75-1, 0.519)};
            
            \end{axis}
        \end{tikzpicture}
        \caption{ }
        \label{fig:minority_percent}
    \end{subfigure}
    \hspace{1cm}
    \begin{subfigure}{0.4\textwidth}
        \centering
        \begin{tikzpicture}
        \begin{axis}[
            ybar,
            bar width=0.2cm,
            width=7cm,
            height=4.5cm,
            legend style={at={(0.75,1)}, anchor=north, legend columns=-1, font=\tiny},
            symbolic x coords={0-0.25, 0.25-0.5, 0.5-0.75, 0.75-1},
            xtick=data,
            xlabel={Loss Quantile},
            ylabel={$p(\text{quantile}|\text{majority})$},
            ymin=0,
            ymax=0.4,
            nodes near coords={},
            tick label style={font=\fontsize{7}{9}\selectfont}
        ]
            \addlegendentry{Waterbirds}
            \addlegendentry{Dominos}

            % Waterbirds
            \addplot[fill=purple] coordinates {(0-0.25, 0.262) (0.25-0.5, 0.257) (0.5-0.75, 0.251)  (0.75-1, 0.23)};
            
            % Dominos 
            \addplot[fill=pink] coordinates {(0-0.25, 0.264) (0.25-0.5, 0.26) (0.5-0.75, 0.255) (0.75-1, 0.22)};
            
            \end{axis}
        \end{tikzpicture}
        \caption{ }
        \label{fig:majority_percent}
    \end{subfigure}
    \caption{(a) Distribution of training images from minority groups between loss quantiles for the Waterbirds and Dominoes datasets. (b) Distribution of training images from majority groups between loss quantiles for the Waterbirds and Dominoes datasets.}
    \label{fig:datapoint_distribution}
\end{figure}

In \cref{sec:heatmap}, we claimed that the images on which the model trained with ERM has a low loss show specific properties. This assumption is valid since on the images from the majority groups both the causal and non-causal parts of images are in accordance with the label. Hence, even if the model attends more to the non-causal parts or its attention is divided between the causal and non-causal parts, it will still perform well on the datapoint and obtains a low loss. \cref{fig:minority_percent} illustrates that the images from minority groups are more among the images with high losses. On the other hand, images from majority groups are almost uniformly distributed between loss quantiles, with a slightly higher probability in lower loss quantiles, as shown in \cref{fig:majority_percent}. Since the probability of majority samples is higher than the minority ones across the dataset  and $p(\text{low loss}|\text{majority})$ is high, it can be concluded that the probability of a low loss sample being from the majority groups is relatively high. 

\section{Comparison of Attribution Maps }
\label{sec:vis}
\begin{figure}[ht!]
\centering
\includegraphics [width = 0.4 \textwidth]{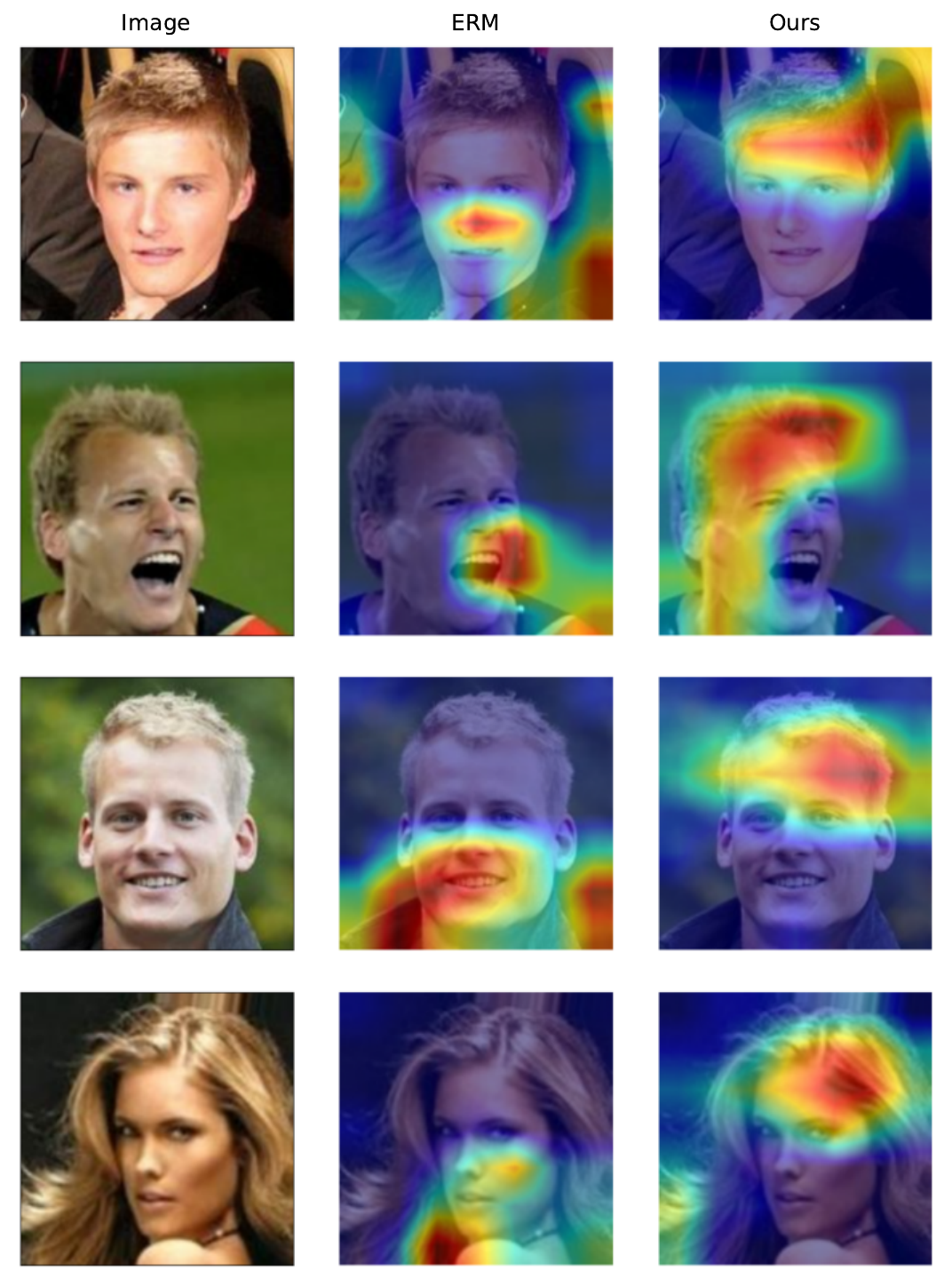}
\caption{Saliency maps of models trained with ERM and our proposed method on CelebA samples which are misclassified by the base model trained with ERM.}
\label{fig:celeba_map}
\end{figure}

\begin{figure}[h!]
\centering
\includegraphics [width = 0.4 \textwidth]{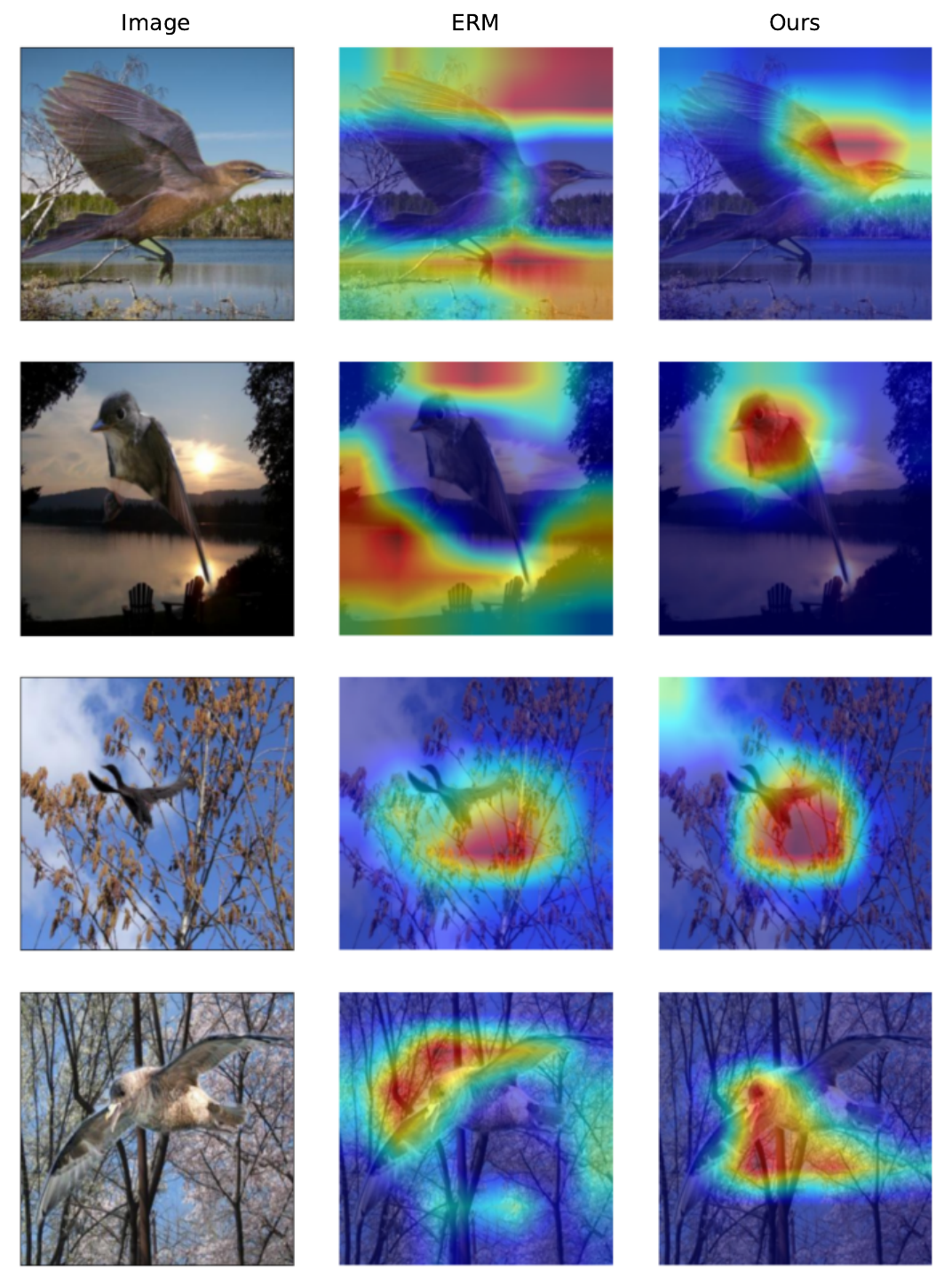}
\caption{Saliency maps of models trained with ERM and our proposed method on Waterbirds samples which are misclassified by the base model trained with ERM.}
\label{fig:wb_map}
\end{figure}

\begin{figure}[h!]
\centering
\includegraphics [width = 0.4 \textwidth]{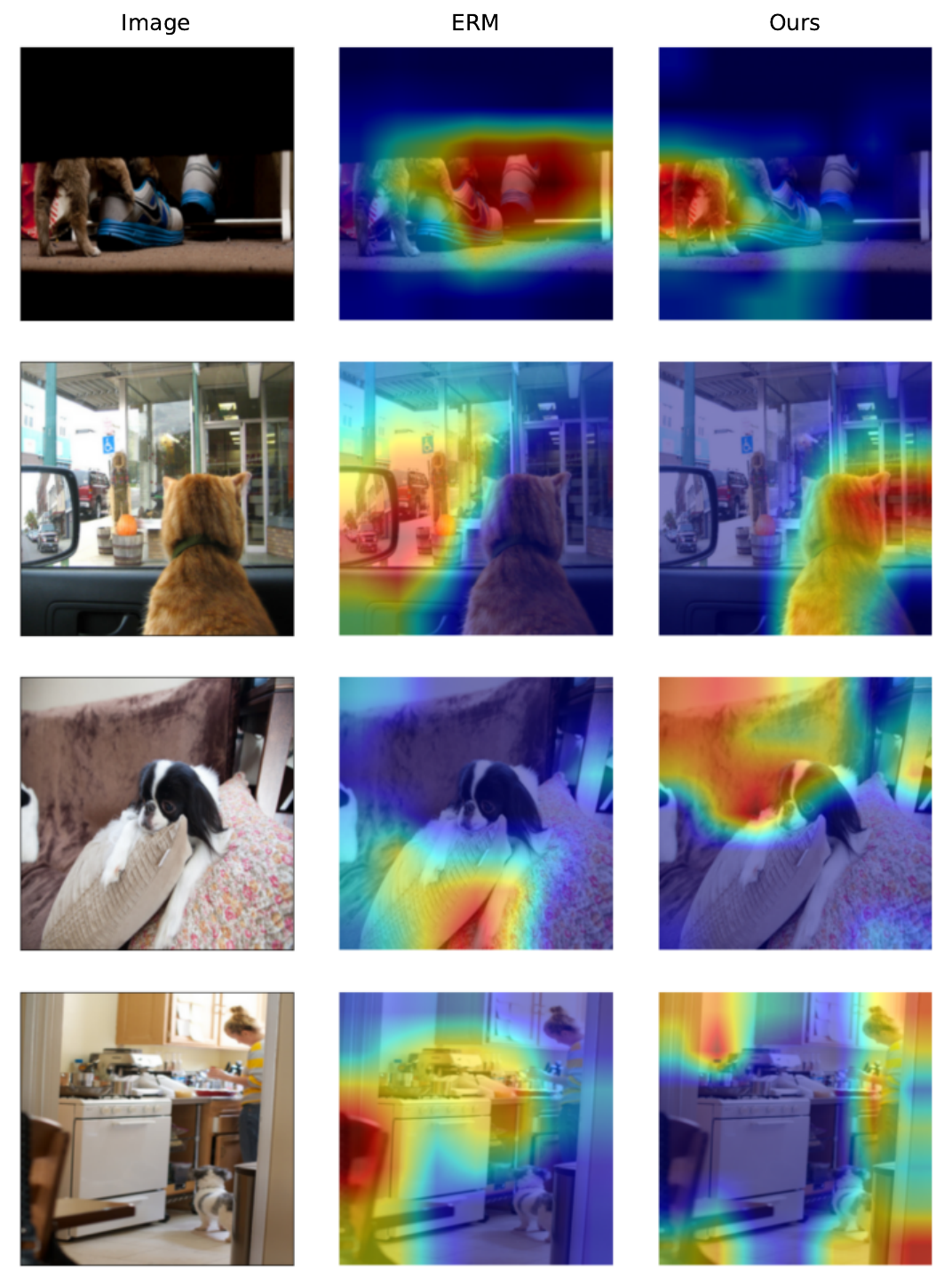}
\caption{Saliency maps of models trained with ERM and our proposed method on Metashift samples which are misclassified by the base model trained with ERM.}
\label{fig:metashift_map}
\end{figure}

\begin{figure}[ht!]
\centering
\includegraphics [width = 0.4 \textwidth]{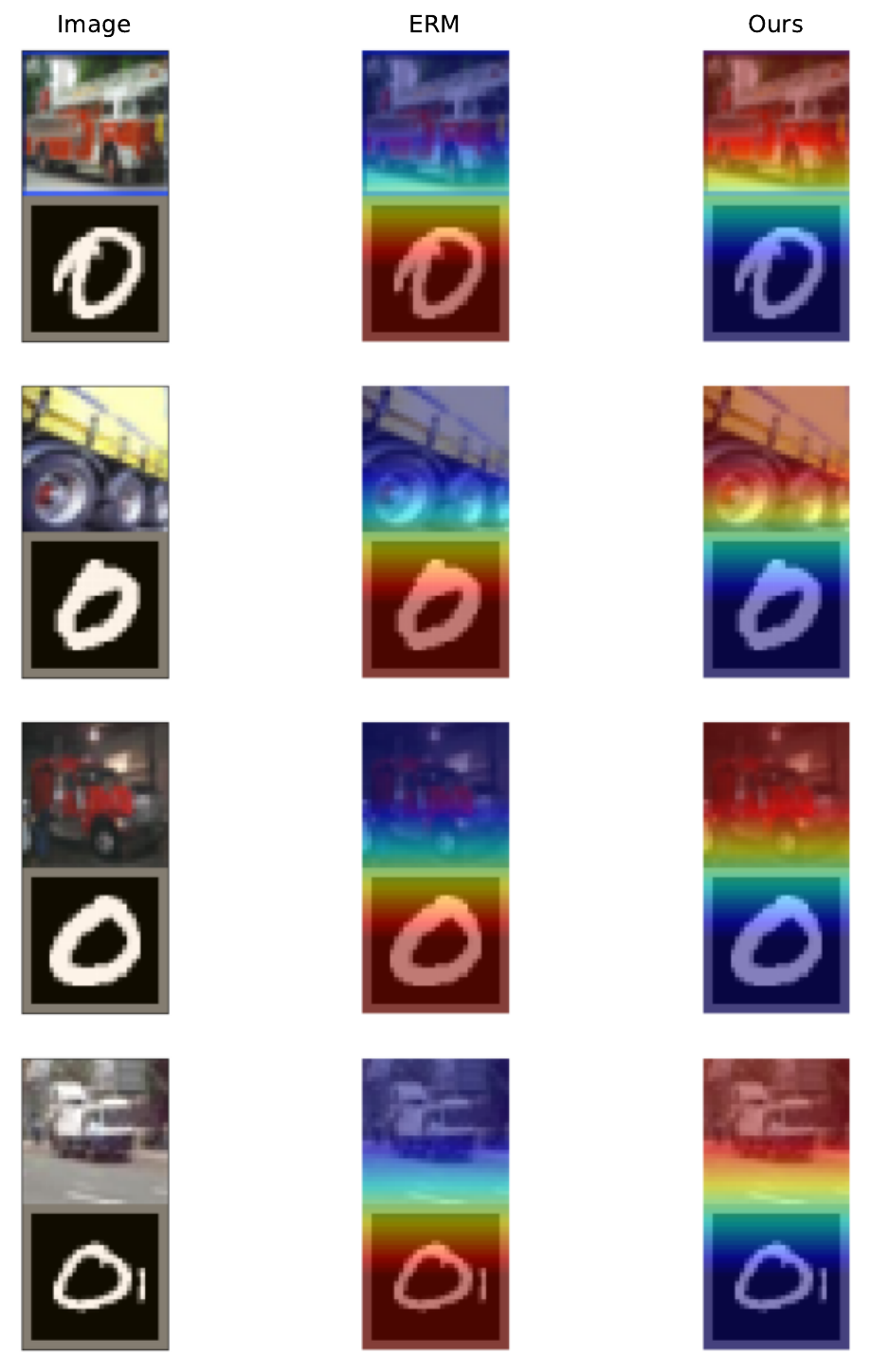}
\caption{Saliency maps of models trained with ERM and our proposed method on Dominoes samples which are misclassified by the base model trained with ERM.}
\label{fig:domino_map}
\end{figure}

The class activation maps of models trained with ERM and our method on some samples are illustrated in \cref{fig:wb_map,fig:celeba_map,fig:metashift_map,fig:domino_map}.

\section{Combined Images}
\begin{figure}[ht!]
\centering
\includegraphics [width = 0.4 \textwidth]{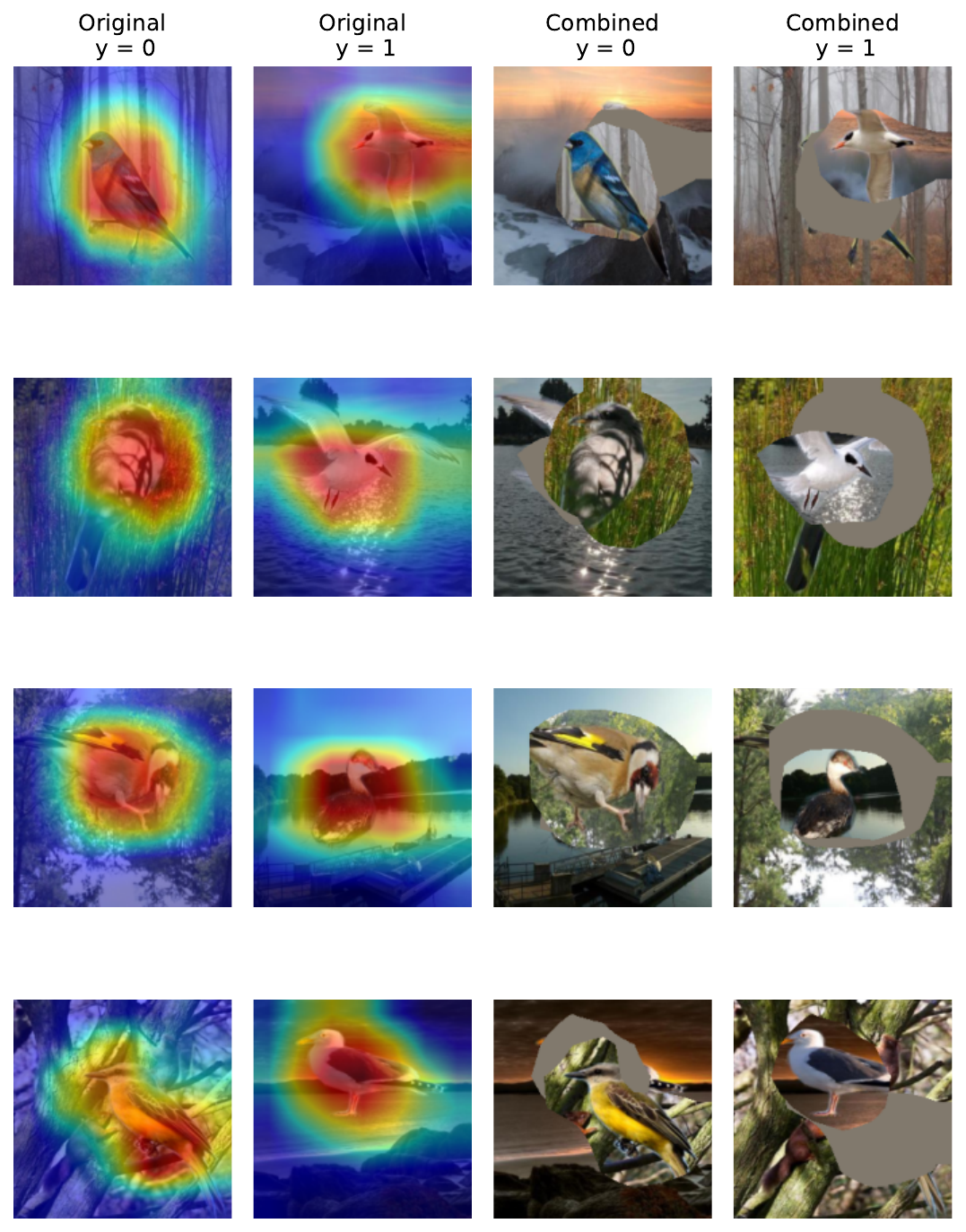}
\caption{Low loss training samples in the Waterbirds and their combinations.}
\label{fig:wb_comb}
\end{figure}

\begin{figure}[ht!]
\centering
\includegraphics [width = 0.4 \textwidth]{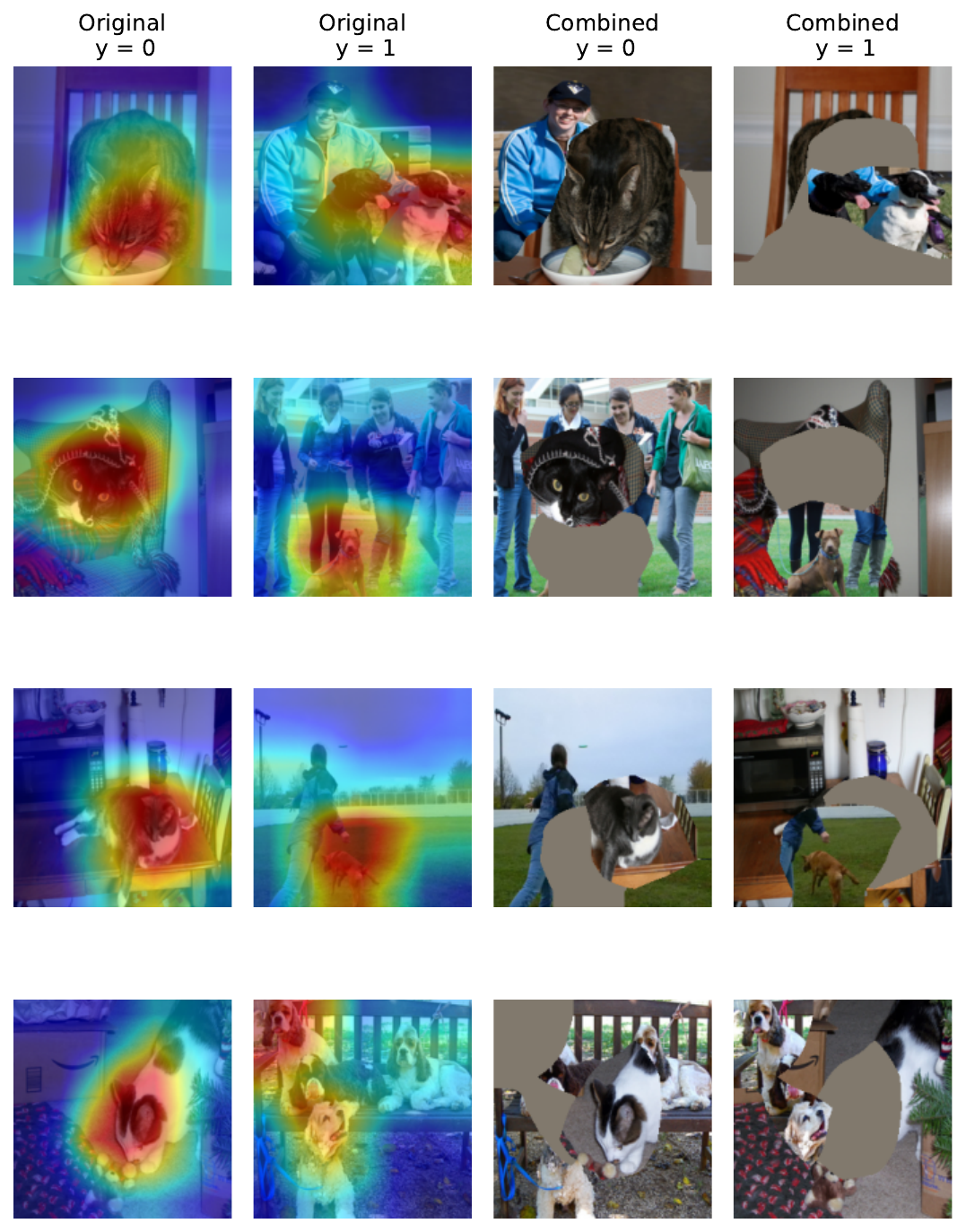}
\caption{Low loss training samples in the Metashift and their combinations.}
\label{fig:metashift_comb}
\end{figure}

\begin{figure}[ht!]
\centering
\includegraphics [width = 0.4 \textwidth]{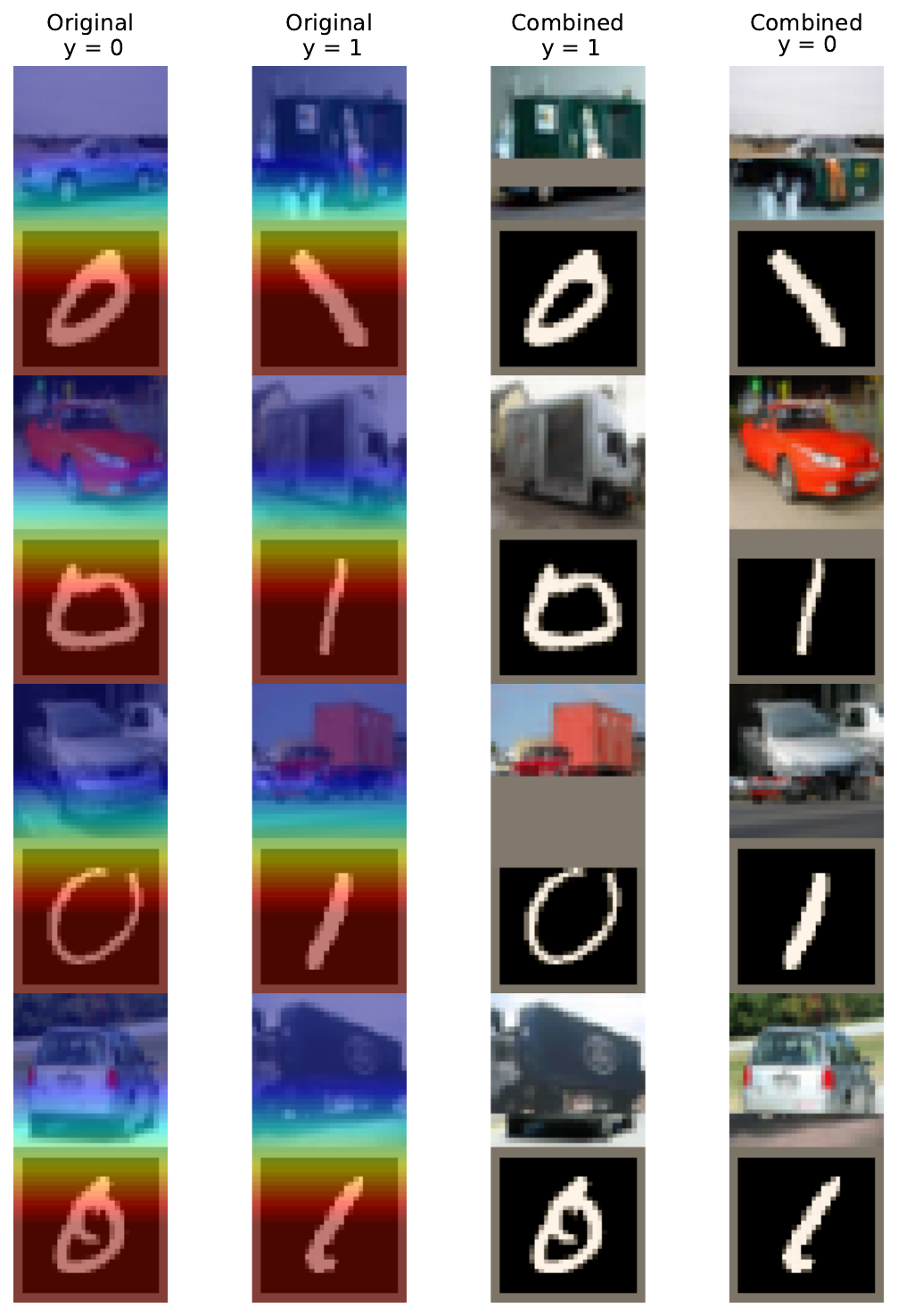}
\caption{Low loss training samples in the Dominoes and their combinations.}
\label{fig:domino_comb}
\end{figure}

Some examples of combined images and their corresponding label are shown in \cref{fig:wb_comb,fig:metashift_comb,fig:domino_comb}.

% To split the supplementary pages from the main paper, you can use \href{https://support.apple.com/en-ca/guide/preview/prvw11793/mac#:~:text=Delete%20a%20page%20from%20a,or%20choose%20Edit%20%3E%20Delete).}{Preview (on macOS)}, \href{https://www.adobe.com/acrobat/how-to/delete-pages-from-pdf.html#:~:text=Choose%20%E2%80%9CTools%E2%80%9D%20%3E%20%E2%80%9COrganize,or%20pages%20from%20the%20file.}{Adobe Acrobat} (on all OSs), as well as \href{https://superuser.com/questions/517986/is-it-possible-to-delete-some-pages-of-a-pdf-document}{command line tools}.


\end{document}